\documentclass{article} 
\usepackage{iclr2026_conference,times}

\usepackage{amsmath,amsfonts,bm}

\def\eqref#1{equation~\ref{#1}}

\def\1{\bm{1}}

\DeclareMathAlphabet{\mathsfit}{\encodingdefault}{\sfdefault}{m}{sl}
\SetMathAlphabet{\mathsfit}{bold}{\encodingdefault}{\sfdefault}{bx}{n}

\usepackage[table]{xcolor}  
\usepackage{xcolor}
\usepackage{hyperref}
\usepackage{url}
\usepackage{xcolor} 
\usepackage{subcaption}
\usepackage{graphicx}
\usepackage{wrapfig} 

\usepackage{xcolor}  
\usepackage{xcolor}
\newcommand{\gnum}[1]{\textcolor{gray}{#1}}
\usepackage{hyperref} 

\title{NAIPv2: Debiased Pairwise Learning for Efficient Paper Quality Estimation}

\author{
Penghai Zhao\textsuperscript{1,\dag}, 
Jinyu Tian\textsuperscript{1,\dag}, 
Qinghua Xing\textsuperscript{1,\dag}, 
Xin Zhang\textsuperscript{1}, 
Zheng Li\textsuperscript{1}, 
Jianjun Qian\textsuperscript{3}, \\
\textbf{Ming-Ming Cheng\textsuperscript{2,1}, 
Xiang Li\textsuperscript{2,1,*}} \\
\textsuperscript{1}VCIP, CS, Nankai University \\
\textsuperscript{2}NKIARI, Shenzhen Futian \\
\textsuperscript{3}PCALab, School of Computer Science and Engineering, \\
Nanjing University of Science and Technology \\
\dag\ Equal contribution \quad * Corresponding authors
}

\iclrfinalcopy 
\begin{document}

\maketitle

\begin{abstract}
The ability to estimate the quality of scientific papers is central to how both humans and AI systems will advance scientific knowledge in the future. However, existing LLM-based estimation methods suffer from high inference cost, whereas the faster direct score regression approach is limited by scale inconsistencies. We present NAIPv2, a debiased and efficient framework for paper quality estimation. NAIPv2 employs pairwise learning within domain-year groups to reduce inconsistencies in reviewer ratings and introduces the Review Tendency Signal (RTS) as a probabilistic integration of reviewer scores and confidences. To support training and evaluation, we further construct NAIDv2, a large-scale dataset of 24,276 ICLR submissions enriched with metadata and detailed structured content. Trained on pairwise comparisons but enabling efficient pointwise prediction at deployment, NAIPv2 achieves state-of-the-art performance (78.2\% AUC, 0.432 Spearman), while maintaining scalable, linear-time efficiency at inference. Notably, on unseen NeurIPS submissions, it further demonstrates strong generalization, with predicted scores increasing consistently across decision categories from Rejected to Oral. These findings establish NAIPv2 as a debiased and scalable framework for automated paper quality estimation, marking a step toward future scientific intelligence systems.
\end{abstract}

\begin{center}
\includegraphics[height=1.0em]{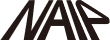} \;
\textbf{Homepage:} \\
\url{sway.cloud.microsoft/Pr42npP80MfPhvj8} \\[0.6em]
\includegraphics[height=1.5em]{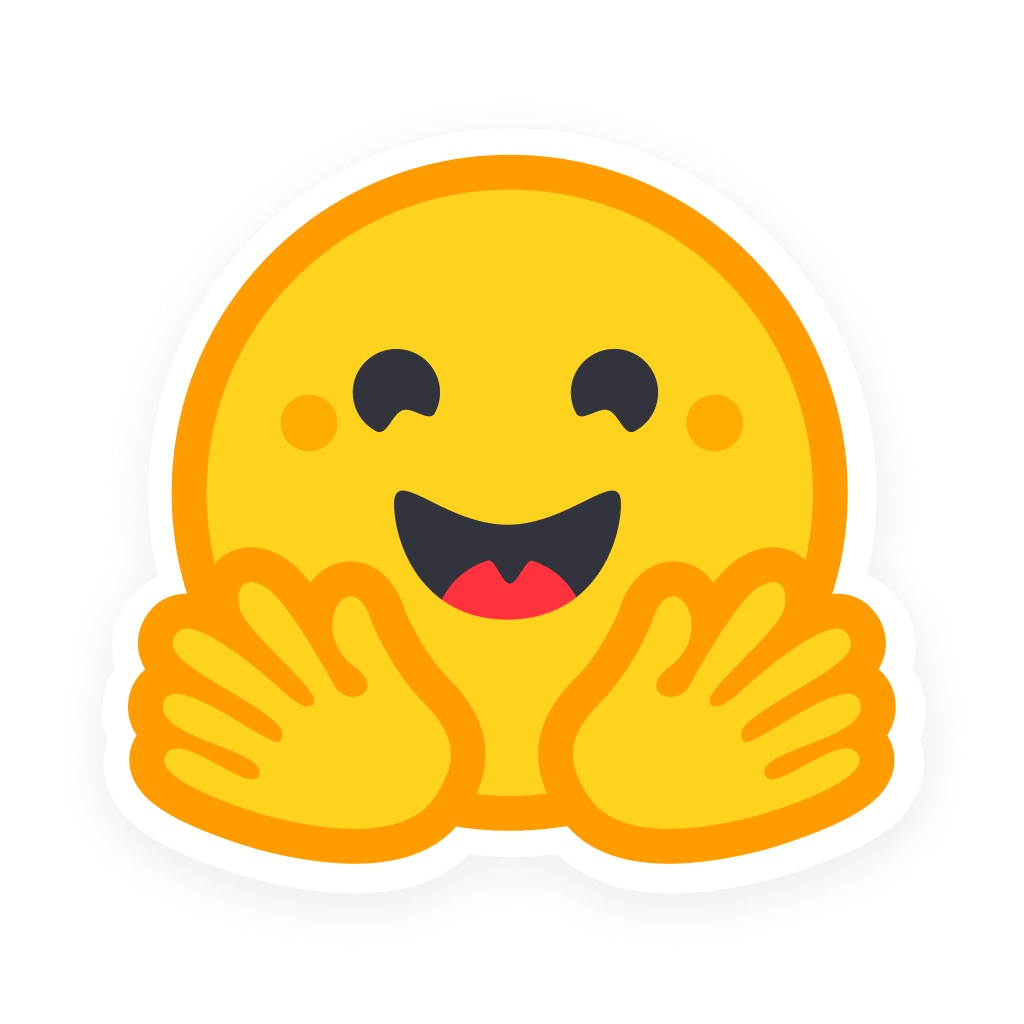} \;
\textbf{Online Demo:} \\
\url{huggingface.co/spaces/ssocean/Newborn_Article_Impact_Predict}
\end{center}

\section{Introduction}

The advent of large language models (LLMs) has propelled scientific research into an unprecedented phase of acceleration. One of the most immediate manifestations is the explosive growth of scholarly publications, with about 242,740 AI-related publications in 2023~\citep{maslej2025artificial}. Such an overwhelming influx of literature not only far exceeds the limits of human reading capacity but also creates significant challenges for automated scientific systems~\citep{lu2024ai,si2024can,baek2024researchagent,wang2024autosurvey,zhao2024literature}. To discern valuable insights and maintain a leading academic perspective, the necessity of focusing on high-quality work in the most recent publications has become increasingly paramount. However, assessing the quality of early-stage articles poses particular challenges, as the absence of citation histories and other bibliometric indicators provides little evidence of their quality. At the same time, both DoRA~\citep{american2012san} and the Leiden Manifesto~\citep{hicks2015bibliometrics} strongly discourage the use of journal or conference prestige as a proxy for paper quality. Consequently, growing attention is directed toward approaches that evaluate research quality based on content-related features rather than external attributes.

\begin{figure}[ht]
\centering
\includegraphics[width=1.0\textwidth]{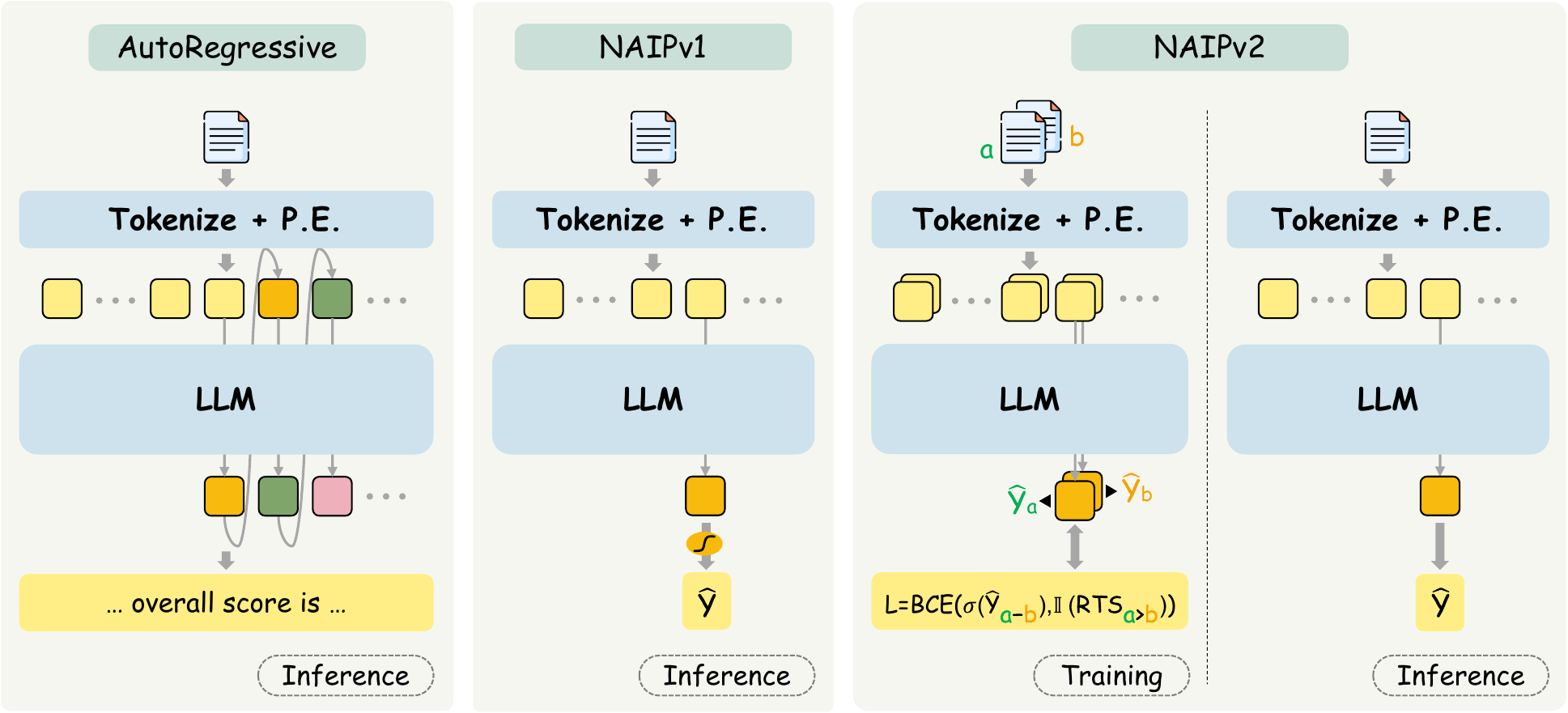} 
\caption{Comparison of various frameworks. AutoRegressive approaches~\citep{zhu-2025-deepreview, weng2025cycleresearcher} rely on sequential generation, resulting in substantial inference latency; NAIPv1~\citep{zhao2025words} enables fast pointwise regression but suffers from scale inconsistency; NAIPv2 leverages debiased pairwise training with confidence-aware signals and operates as an efficient pointwise regressor with linear-time complexity at inference.}
\label{fig:framework}
\end{figure}

For decades, expert peer review has served as a reliable standard for evaluating research quality . While rigorous, this process is costly, time-consuming, and inherently difficult to scale. In response, recent studies~\citep{yu2024automated,zhu-2025-deepreview,d2024marg,idahl2024openreviewer} have investigated the use of LLMs to assist in the review process (\textit{e.g.,} generating review comments, predicting review scores, or inferring acceptance decisions directly from paper content). As shown in Fig.~\ref{fig:framework} left panel, while these approaches demonstrate promising potential, their reliance on autoregressive reasoning over long contexts introduces substantial latency and computational overhead, thereby limiting deployment in automated scientific systems (\textit{e.g.}, literature intelligence systems).

To mitigate the inherent latency of autoregressive models, we re-examine regression approaches that target citation-based indicators (middle part of Fig.~\ref{fig:framework}). Despite recent research exploring the direct prediction of various citation-based indicators from textual content~\citep{zhao2025words,de2024can}, such approaches primarily emphasize forecasting research impact rather than assessing the intrinsic quality of papers. As reported by ~\cite{cortes2021inconsistency} and ~\cite{dougherty2022citation}, citation counts are considered unreliable proxies of article quality. This underscores the need to shift quality estimation toward more intrinsically reliable signals (\textit{i.e.}, peer review score). However, our experiments reveal that directly regressing on review scores remains unsatisfactory. We attribute this shortcoming to two fundamental factors: First, review scores may lack a consistent and stable measurement scale. The distribution of review scores differs across research domains and publication years. In addition, conference evaluation standards and invited expert reviewers may shift over time. Second, reviewer confidence is largely neglected in existing formulations. Each review score is typically accompanied by a confidence rating, reflecting the reviewer’s self-assessed reliability of judgment. Nevertheless, most approaches ignore this signal, treating all scores as equally trustworthy. This neglect allows low-confidence assessments to exert the same influence as high-confidence ones, introducing additional label noise and weakening the supervisory signal.

Building on these observations, we propose NAIPv2 (originating from Newborn Article Impact Prediction), a fast and debiased framework for paper quality assessment. NAIPv2 learns the relative ordering of submissions within the same domain–year group in a pairwise manner, thereby mitigating biases introduced by cross-domain, temporal variations. During inference, the framework is decoupled into a simplified single-branch pointwise regressor, enabling robust score prediction while preserving linear-time complexity. In addition, we introduce a probabilistic treatment of review scores and reviewer confidences, where each score is viewed as a noisy observation of the latent paper quality and confidence determines its uncertainty, yielding a principled aggregation signal. Experimental results demonstrate that our method attains state-of-the-art performance (78.2\% AUC and 0.432 Spearman) while offering substantially improved inference efficiency. Notably, on unseen NeurIPS submissions, the predicted scores align precisely with decision categories, consistently assigning the lowest values to rejected papers and the highest values to oral presentations.

In summary, our main contributions are as follows:

\begin{enumerate}

    \item \textbf{A large-scale domain-labeled dataset for quality estimation.} We release \textbf{NAIDv2}, a large-scale dataset of 24,276 ICLR submissions (2021–2025) that combines metadata, parsed PDF content, and clustered labels to enable cross-domain debiasing and foster research on automated peer review and scholarly evaluation.  

    \item \textbf{A scalable and debiased framework for scientific article quality estimation.} We propose a scalable and debiased framework \textbf{NAIPv2} for early-stage paper quality prediction. Beyond achieving state-of-the-art performance, NAIPv2 exhibits strong generalization and consistency on unseen NeurIPS review data.

    \item \textbf{Confidence-aware probabilistic signal.} We introduce the \textbf{RTS}, a probabilistic formulation that models review scores as Gaussian observations of latent paper quality with reviewer confidence shaping uncertainty, providing a principled aggregation that yields a reliable comparative signal and clarifies the role of confidence in evaluation outcomes.  

\end{enumerate}

\section{Related Work}

\textbf{Peer Review Dataset.}
With its open-access policy, ICLR has become a widely used primary data source in contemporary research~\cite{staudinger2024analysis}. However, such reliance also introduces considerable risks of information leakage. The first type of leakage is induced by large language models. For example, PeerRead~\citep{kang2018dataset} is an early and influential dataset, containing more than 14K paper drafts with accept/reject decisions from leading venues. Given its early release, the dataset has likely been incorporated into LLM training corpora, thereby raising concerns about unintended information leakage. The second type of leakage results from overlaps between training and test sets. Recently,~\cite{zhang2025re} introduced the Re$^{2}$ dataset. However, some of its benchmark data may overlap with training sets employed by other methods, which can lead to deviations between reported results and those presented in the original papers. Following rigorous academic practice, we construct the NAIDv2 test set exclusively from the latest ICLR 2025 data, thereby minimizing the risk of potential information leakage.

\textbf{Automated Peer Review.} 
To optimize performance on review-related tasks, most automated peer-review methods construct dedicated datasets and perform task-specific fine-tuning. ~\cite{yu2024automated} introduces a modular framework for standardization, evaluation, and analysis, and further proposes a mismatch score to measure paper–review alignment. OpenReviewer~\citep{idahl2024openreviewer} fine-tunes an 8B-parameter LLM on expert reviews to produce structured, guideline-compliant feedback. DeepReview~\citep{zhu-2025-deepreview} employs a multi-stage pipeline integrating structured analysis, literature retrieval, and evidence-based reasoning. CycleReviewer~\cite{weng2025cycleresearcher} goes beyond single-pass reviewing by coupling manuscript generation with iterative peer review through reinforcement learning, aiming to automate the entire research-review cycle. Together, these approaches exploit LLMs’ CoT reasoning, boosting performance but incurring substantial computational cost. Instead, NAIPv2 adopts pairwise learning to address scale inconsistency and decouples into a pointwise regressor at inference, maintaining competitive accuracy with markedly improved speed.



\textbf{Numerical Impact and Quality Estimation.}
Predicting the future impact of scientific papers has long been an active area of study~\citep{zhao2022utilizing,xia2023review,thelwall2025research}, typically leveraging early signals such as citation counts, author profiles, venue prestige, and linguistic features. More recently, LLM-based approaches have been explored: ~\cite{de2024can} investigated whether ChatGPT can estimate bibliometric indicators including citation counts, readership, and social media interactions, finding stronger correlations than traditional readability metrics, while ~\cite{zhao2025words} employed LLMs to directly predict a field- and time-normalized impact score ($TNCSI_{sp}$) from titles and abstracts, achieving an MAE of 0.216. 

A number of methods have also been proposed to estimate research quality. Prior research~\citep{ghosal2019deepsentipeer,bharti2021peerassist} has primarily focused on using machine learning algorithms to predict final recommendation scores or classify papers into accept/reject outcomes. Such practices, however, tend to overlook the role of confidence annotations. In contrast, our approach jointly models scores and confidence in a probabilistic framework, thereby preserving the semantics of uncertainty and enabling principled aggregation across reviewers.

\textbf{Pairwise Ranking.}
Pairwise ranking methods have long been established as effective approaches in information retrieval and recommendation~\citep{liu2009learning,cao2007learning}, yet their adoption in peer review and scientometric prediction remains limited. \cite{hopner2025automatic} apply pairwise ranking for score prediction and find it feasible for citation counts but unreliable for review scores. \cite{zhang2025replication} investigate pairwise manuscript evaluation using off-the-shelf LLMs. Owing to the quadratic growth of pairwise comparisons, they demonstrate that aggregating fewer than 2\% of all possible pairs is sufficient to recover a robust global ranking. Distinct from prior studies, this work addresses debiasing and efficiency by training in a pairwise manner while performing pointwise inference, thereby achieving inference with linear complexity.

\section{Methods}
\subsection{Review Tendency Signal (RTS)}
In practice, reviewers are not always domain experts in the exact area of a submission, which may prevent them from providing accurate evaluations. To mitigate this source of uncertainty, most conferences and journals require reviewers to report not only an overall score but also an accompanying confidence level, reflecting the degree of trust they place in their own assessment.

Our proposed solution is to interpret each review not as a deterministic score but as a noisy observation of the latent “true quality” of the paper. In this view, the reported overall score $s_i$ is treated as the mean of a probability distribution, while the confidence level $c_i$ determines the variance (or equivalently, the precision) of that distribution. High-confidence reviews correspond to narrow distributions centered around their scores, whereas low-confidence reviews correspond to wider distributions, reflecting greater uncertainty.

Formally, we assume each review $i$ provides an observation modeled as a Gaussian likelihood. Here, both the reported score and the confidence level are normalized to the unit interval $[0,1]$.
\begin{equation}
p(s_i \mid x, c_i) \;=\; \mathcal{N}\!\left(s_i \mid x, \sigma(c_i)^2\right),
\end{equation}
where $x \in [0,1]$ denotes the latent true quality, and $\sigma(c_i)$ is a decreasing function of confidence, ensuring that higher confidence yields lower variance.

Given $n$ independent reviews, the posterior distribution over $x$ can be written as:
\begin{equation}
p(x \mid s_{1:n}, c_{1:n}) \;\propto\; \prod_{i=1}^n p(s_i \mid x, c_i).
\end{equation}
We then define the Review Tendency Signal (RTS) of a submission as the posterior mean:
\begin{equation}
\mathrm{RTS} \;=\; \mathbb{E}[x \mid s_{1:n}, c_{1:n}].
\end{equation}

This probabilistic formulation naturally balances reviewer scores according to their associated uncertainties. Unlike weighted averaging, which collapses confidence to a fixed coefficient, $RTS$ preserves the semantics of uncertainty, enables principled aggregation across reviewers, and guarantees that the resulting estimate lies within the valid range $[0,1]$. Defination of $\sigma(c_i)$ and more details are provided in the Appendix~\ref{app:RTS_norm}.

\subsection{Construction of the NAIDv2 Dataset}

To facilitate efficient debiasing in training and evaluation, we construct the NAIDv2 (where D denotes Dataset). Following established community practices, we build it from ICLR conference review data provided by OpenReview.net, ensuring an acceptance ratio that closely reflects real-world conditions. Given that review data is continually updated, earlier studies either failed to incorporate the latest ICLR reviews or covered a narrower time span~\citep{zhu-2025-deepreview, weng2025cycleresearcher}. To address this, we developed a dedicated data engine and used it to compile NAIDv2, encompassing 24,276 samples spanning 2021–2025. Although our experiments relied only on review scores, confidence ratings, and related fields, the underlying SQL-based dataset also retains additional information, including the full textual review comments. Furthermore, we downloaded the corresponding original PDF submissions and applied MinerU for structured parsing.

\begin{figure}[h]
\centering
\includegraphics[width=1.0\textwidth]{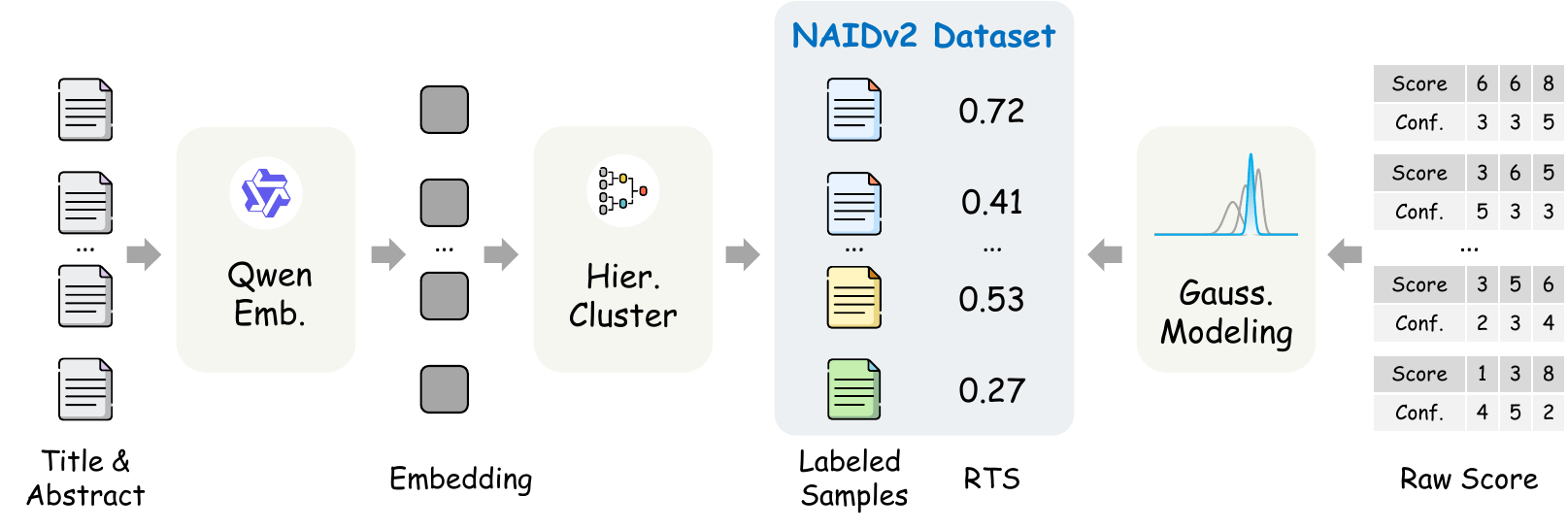} 
\caption{\textbf{Debiased data construction in NAIDv2.} Titles and abstracts are embedded and hierarchically clustered into latent domains to mitigate domain bias, while raw scores with reviewer confidences are modeled via Gaussian likelihoods to calculate RTS, yielding a normalized dataset for robust debiased pairwise training.}
\label{fig:naid_pipeline}
\end{figure}

In addition to its scale and coverage, NAIDv2 is distinguished by its explicit handling of domain bias (as depicted in Fig.~\ref{fig:naid_pipeline}). As discussed earlier, raw review scores exhibit inconsistencies across domains and across time. To correct for these distortions, we move beyond noisy keyword-based domain labeling and instead adopt a clustering-driven strategy. Specifically, we encode paper titles and abstracts using Qwen3-Embedding-4B~\citep{zhang2025qwen3} and apply hierarchical clustering~\citep{bar2001fast,mullner2011modern} to identify latent domains based on semantic relatedness. Accordingly, we also compute the RTS from the raw scores, providing a robust signal for debiased training. This pipeline produces domain-normalized labels that more faithfully capture expert consensus and enable debiased pairwise training.

For additional information on dataset construction, please refer to the Appendix~\ref{app:NAIDv2 Engine}.

\subsection{Pairwise Training and Pointwise Scoring with NAIPv2}

\textbf{Pairwise Training Stage}.
In the training phase, NAIPv2 adopts a pairwise learning paradigm, where the model is optimized to capture relative quality differences between two papers rather than directly regressing absolute scores (depicted in the rightmost panel of Fig.~\ref{fig:framework}). 
Given a pair of submissions $(a, b)$ with ground-truth review tendency signal ${RTS}_a$ and ${RTS}_b$, we define a binary label ${RTS}_{ab} = \mathbb{I}[\,{RTS}_a > {RTS}_b\,]$ and obtain predictions $\hat{y}_a, \; \hat{y}_b$ from a shared LLaMA-3~\citep{dubey2024llama} backbone. $\mathbb{I}[\cdot]$ denotes the indicator function, which equals $1$ if the condition holds and $0$ otherwise. The pairwise preference probability is:
\begin{equation}
\hat{z}_{ab} = sigmoid(\hat{y}_a - \hat{y}_b).
\label{eq:pairwise-prob}
\end{equation}
The training objective is the binary cross-entropy loss.
\begin{equation}
\mathcal{L}(a,b) = -\Big( {RTS}_{ab}\,\log \hat{z}_{ab} + (1-{RTS}_{ab})\,\log(1-\hat{z}_{ab}) \Big).
\label{eq:pairwise-loss}
\end{equation}

To mitigate distributional biases across domains and years, training pairs $(a, b)$ are restricted to submissions within the same cluster and the same publication year. This design prevents spurious comparisons caused by cross-field or temporal inconsistencies. 

To further enhance robustness and stability, we employ a curriculum learning mechanism based on the gap in $RTS$. Specifically, we partition training pairs into difficulty buckets according to:
\begin{equation}
\Delta_{ab} = |\,{RTS}_a - {RTS}_b\,|,
\label{eq:rts-gap}
\end{equation}
where pairs with larger $\Delta_{ab}$ are considered easier, while those with smaller $\Delta_{ab}$ are treated as harder. During the early stages of training, easier pairs dominate the sampling distribution, guiding the network to quickly capture coarse quality distinctions. As training progresses, harder pairs are gradually upsampled, enabling the model to learn fine-grained quality differences.

\textbf{Pointwise Inference Stage}.
Although training supervision is provided only on submission pairs $(a,b)$, the shared backbone guarantees that each input is independently mapped to a scalar score:
\begin{equation}
\hat{y}_a = f(x_a;\theta), \quad \hat{y}_b = f(x_b;\theta),
\label{eq:pointwise-mapping}
\end{equation}
where $x_a$ and $x_b$ denote the textual representations of submissions $a$ and $b$, respectively. 
In other words, the pairwise objective implicitly induces a decoupled pointwise scoring function under shared parameters $\theta$. 

At inference time, this function naturally generalizes to single-paper evaluation:
\begin{equation}
\hat{y} = f(x;\theta),
\label{eq:pointwise-infer}
\end{equation}
which provides a pointwise estimate of paper quality that is directly comparable across submissions within the same domain-year subgroup. 
Although the pairwise loss constrains only relative differences, the shared backbone forces all submissions into a common representation space, thereby yielding in practice a globally consistent scoring scale (up to an additive constant).

\section{Experiments}
\subsection{Metrics}
We evaluate the performance of our method from two different viewpoints. 
The first viewpoint focuses on predicting whether a paper is accepted or rejected, an interesting and practical task evaluated by classification accuracy.
The second viewpoint addresses the requirements of literature intelligence systems (e.g., recommendation systems, automated research assistants, or automatic literature review generation system), where the emphasis is on the quality of score-based ranking and recommendation.

For classification evaluation, we report Accuracy (Acc), F1-score (F1), ROC-AUC (AUC), and Pairwise Accuracy (P. Acc). AUC is defined as $\mathrm{AUC} = 1/(|P||N|)\sum_{i \in P}\sum_{j \in N}\mathbb{I}[\hat{y}_i > \hat{y}_j]$, where $P$ and $N$ are the positive and negative sets. PairAcc measures pairwise directional correctness using thresholded predictions as $\mathrm{PairAcc} = 1/|\Omega|\sum_{(i,j)\in\Omega}\mathbb{I}[(\tilde{y}_i-\tilde{y}_j)(y_i-y_j)>0]$, where $\Omega=\{(i,j): i<j,\; y_i\ne y_j\}$. For regression-based methods, classification results are reported at the threshold yielding the best F1 score.

For ranking evaluation, we use Spearman’s $\rho$ and NDCG@$k$. Spearman’s $\rho$ is $\rho = 1 - 6 \sum_{i=1}^n d_i^2 / (n(n^2 - 1))$ where $d_i$ is the rank difference. NDCG@$k$ is $\mathrm{NDCG@}k = (1/\mathrm{IDCG@}k) \sum_{i=1}^k (2^{s_i}-1)/\log_2(i+1)$ where $s_i$ is the ground-truth relevance and $\mathrm{IDCG@}k$ is the maximum DCG. We set $k=20$ in experiments.

\subsection{Implementation Details}

We divide NAIDv2 into training, validation, and test subsets. 
The validation set is sampled from the training data with an internal 9:1 split and is used for hyperparameter tuning. 
To avoid potential information leakage (e.g., earlier ICLR submissions that might have been included in the LLaMA-3 pretraining corpus), the test set is restricted to submissions from the year 2025, comprising 1,029 samples. NAIPv2 is fine-tuned on 4$\times$A40 GPUs (48\,GB each) using 8-bit quantization and LoRA adaptation~\citep{hu2022lora}, with a training time of roughly 1 hour for 10k pairs (constructed from 23,247 samples). On a more common consumer GPU (RTX~3090, 24\,GB), the same configuration is conservatively estimated to complete within 6 hours, making the approach accessible to most academic labs. The main hyperparameters include a learning rate of $1\times10^{-4}$, a batch size of 8, and training for one epoch.

\subsection{Comparison with Previous Approaches}

As shown in Tab.~\ref{tab:comparison_sota}, we compare our proposed method against various approaches. Existing methods can be broadly divided into three categories. Category I consists of prompting LLMs to predict review scores or acceptance decisions based on paper titles, abstracts, or full texts. Owing to the training-free nature of API calls, this paradigm has gained popularity, though its long-term usage costs remain high. Category II encompasses approaches that fine-tune LLMs for automatic reviewing tasks or develop specialized frameworks. The goal of these methods is to maximize the reasoning and analytical strengths of LLMs, enabling them to produce comprehensive review texts. Category III fine-tunes LLMs to regress specific numerical values. Since these approaches avoid the autoregressive generation process, they achieve much faster inference speeds. In addition, we also report the empirical upper and lower bounds on the NAIDv2 Test Set. The upper bound is obtained by training an MLP with cross-entropy loss, where the inputs are reviewer scores and the ground truth is the binary acceptance decision (0/1). The lower bound is determined by a stochastic model.

\begin{table}[t]
\centering
\caption{Performance comparison on the ICLR review prediction task. $\uparrow$ denotes the higher the better. 
Results marked with * are taken from~\cite{zhu-2025-deepreview,yu2024automated}. ``-'' means the metric is not reported. 
Bold numbers denote the best results.}
\label{tab:comparison_sota}
\begin{tabular}{llccccc}
\hline
\textbf{Category} & \textbf{Method} & \textbf{Acc} $\uparrow$ & \textbf{F1} $\uparrow$ & \textbf{AUC} $\uparrow$ & \textbf{NDCG} $\uparrow$ &  $\boldsymbol{\rho}$ \\
\hline
-   & Lower Bound (Random)     & \gnum{0.514} & \gnum{0.410} & \gnum{0.527} & \gnum{0.525} & \gnum{0.002} \\
    & Upper Bound (Info. Leak) & \gnum{0.819} & \gnum{0.757} & \gnum{0.894} & \gnum{0.995} & \gnum{0.984} \\
\hline

API   & API-pointwise (ChatGPT)    & 0.644 & 0.427 & 0.654 & 0.7024 & 0.315 \\
    & API-pairwise (ChatGPT) & 0.658 & 0.448 & 0.655 & 0.686 & 0.297 \\
    & AgentReview* (Gemini) &  0.678 & 0.559 & - & -& 0.279  \\
    & AI Scientist* (Gemini) &  0.678 & 0.573 & - & -& 0.267  \\
\hline
AutoRegressive  & SEA-EA* (7B)         & 0.582 & 0.563 & - & - & - \\
    & CycleReviewer* (8B) &  0.678 & 0.559 & - & -& 0.279  \\
    & CycleReviewer* (70B) &  0.678 & 0.573 & - & -& 0.267  \\
    & DeepReviewer* (14B) &  0.689 & \textbf{0.623} & - & -& 0.408  \\
\hline
Regress Only & NAIPv1 (8B) & 0.545 & 0.472 & 0.605 & 0.629 & 0.183 \\
    & \textbf{NAIPv2} (\textit{ours}, 8B)    & \textbf{0.706} & 0.609 & \textbf{0.782} & \textbf{0.771}& \textbf{0.432} \\
\hline
\end{tabular}
\end{table}

For the first category, we implement the ``API (pointwise)'' and ``API (pairwise)'' approaches with reference to ~\cite{hopner2025automatic} (see Appendix~\ref{app:API_details} for details). This category has emerged as the most common choice among researchers due to its ease of use; however, neither the pointwise nor the pairwise variant has demonstrated satisfactory performance. The second research direction explores autoregressive inference using LLMs. Due to supervised fine-tuning on specialized data, these methods typically demonstrate stronger performance. DeepReviewer~\citep{zhu-2025-deepreview} reaches an F1 score of 62.3 and a Spearman correlation of 0.408 on their test set, demonstrating strong capability on both acceptance decision prediction and score ranking tasks. Nevertheless, since the method generates tokens autoregressively, inference time scales with output length. On mainstream consumer GPUs (e.g., RTX 3090), inference for a single paper can take up to three minutes, which limits its applicability in literature intelligence systems. Regression-only based approaches seek to bypass autoregression by directly predicting a single numerical output. NAIPv1~\citep{zhao2025words}, for example, achieves up to ten predictions per second on the RTX 3090. However, since its training objective is aligned with influence metrics rather than paper quality, its predictive quality is only marginally better than random guessing. By contrast, our proposed approach leverages a debiased pairwise learning design, matching NAIPv1 in inference efficiency (over a thousand times faster than typical autoregressive approaches) while attaining performance on par with DeepReviewer.

\subsection{Comparison of Learning Paradigms and Inference Complexity}
\begin{wraptable}[10]{r}{0.52\textwidth}  
\centering
\captionof{table}{Comparison of alternative paradigms. The last column reports the theoretical lower-bound inference complexity of each paradigm.}
\label{tab:arch_compare}
\begin{tabular}{lccc}
\hline
\textbf{Paradigm} & \textbf{AUC} & $\boldsymbol{\rho}$ & \textbf{Th. Compl.} \\
\hline
Pointwise & 0.633 & 0.237 & $O(C)$ \\
Pairwise Concat & 0.720 & 0.351 & $O(C \log C)$ \\
NAIPv2 (\textit{ours}) & 0.782 & 0.432 & $O(C)$ \\
\hline
\end{tabular}
\end{wraptable}
p
We further analyze the paradigm differences to underscore the advantages of the NAIPv2 design. Table~\ref{tab:arch_compare} compares three variants: (i) a pointwise method based on the NAIPv1 architecture but using RTS labels in place of the originals, (ii) a pairwise variant that adopts the same architecture as the pointwise model but concatenates two papers into a single input sequence and predicts a binary 0/1 label, and (iii) our proposed NAIPv2. The pointwise approach suffers from inconsistencies, resulting in suboptimal performance. The concatenation-based variant achieves higher accuracy, but its inference requires explicit pairwise comparisons, leading to a complexity of $O(C \log C)$, where $C$ denotes the number of candidate papers. In contrast, NAIPv2 incorporates debiased pairwise learning during training while retaining pointwise inference, thus preserving the linear complexity $O(C)$ of pointwise methods and achieving superior performance.

\subsection{Effect of Clustering Strategies on Performance}
The strategy chosen for pairwise grouping plays a crucial role in the model’s capacity to distinguish between papers within the same set. Table~\ref{tab:debias} presents the performance differences across grouping methods. When no grouping is applied, or when grouping is based solely on temporal information, the model achieves only marginal improvements over direct prediction, and the results remain below optimal. Grouping via GPT-generated keywords performs worst, with accuracy approaching random guessing. This outcome is mainly attributed to the overabundance of candidate keywords, which produces sparse valid pairs and insufficient training data. In contrast, hierarchical clustering outperforms temporal grouping alone, and the combination of temporal and hierarchical grouping yields the strongest overall results. These findings highlight the critical role of debiasing in pairwise learning for paper quality assessment.

\begin{figure*}[t]
\centering
\begin{minipage}{0.48\linewidth}
\centering
\includegraphics[width=\linewidth]{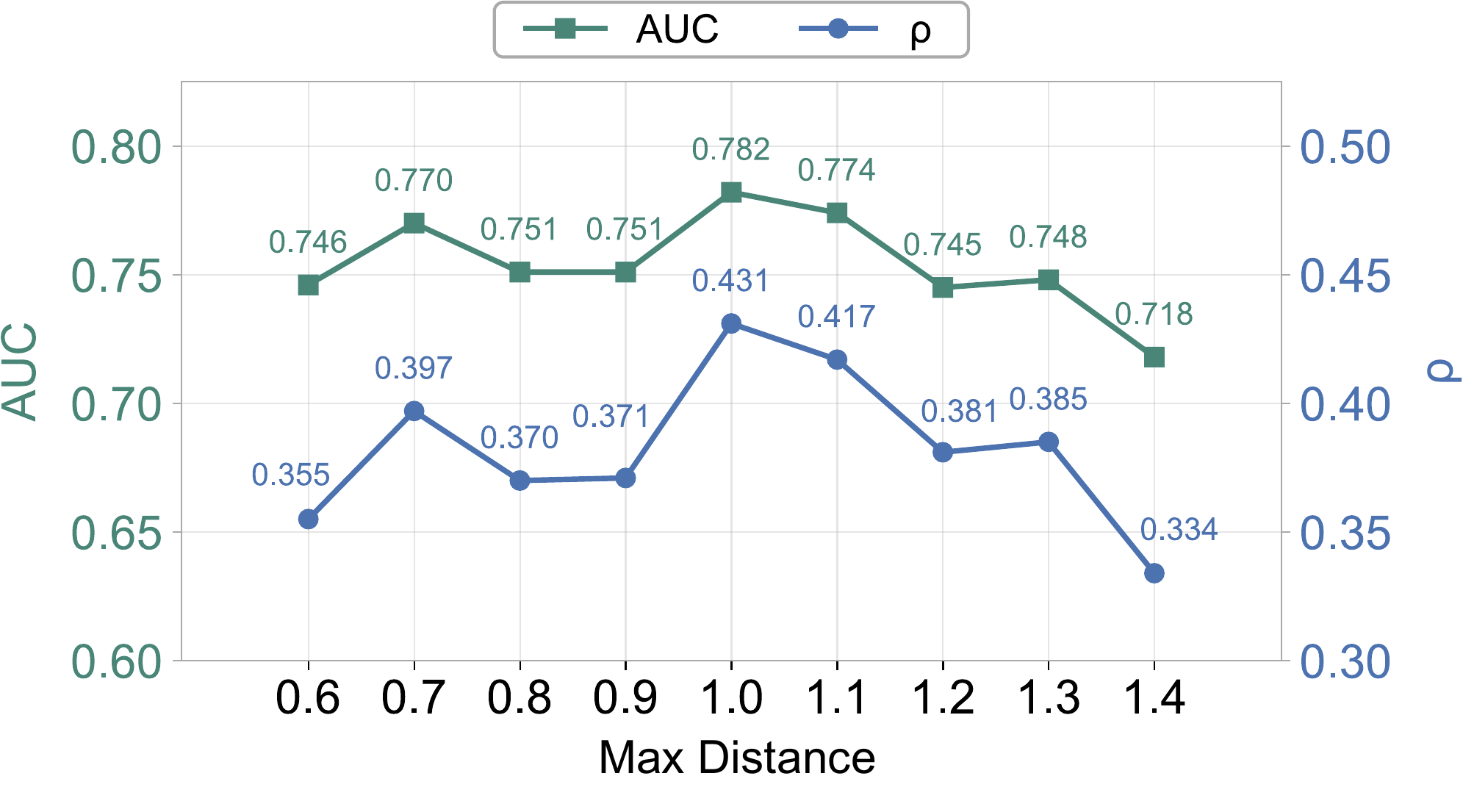}
\caption{Effect of clustering granularity.}
\label{fig:granularity}
\end{minipage}
\hfill
\begin{minipage}{0.48\linewidth}
\centering
\includegraphics[width=\linewidth]{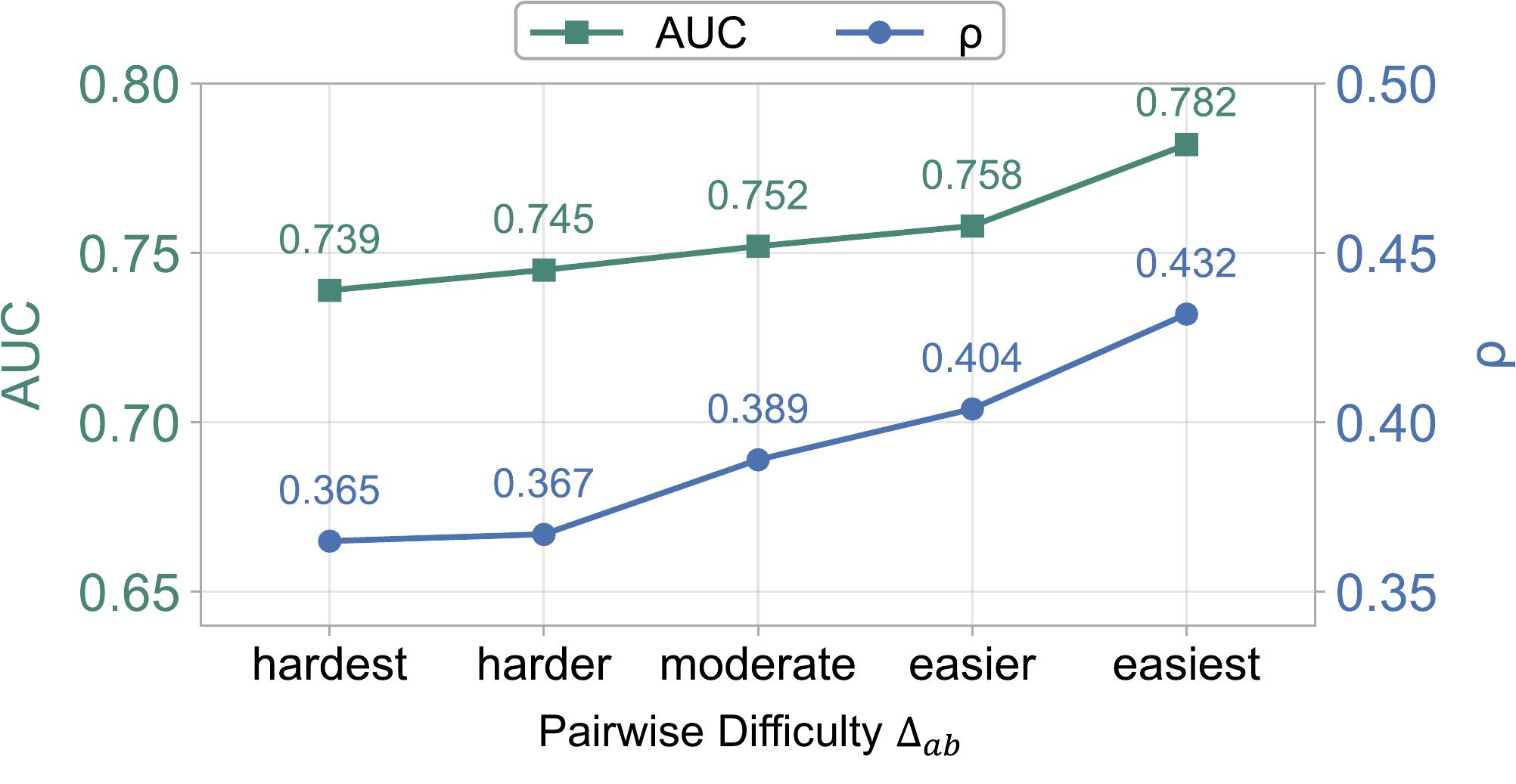}
\caption{Effect of difficulty bucketing.}
\label{fig:curriculum}
\end{minipage}
\end{figure*}

Figure~\ref{fig:granularity} further illustrates the effect of the maximum distance parameter (max distance) in hierarchical clustering. A small max distance produces numerous small clusters with limited samples, thereby suffering from sparsity similar to keyword-based grouping. Conversely, a large max distance leads to a few large clusters, essentially collapsing the structure into a setting akin to no grouping. The results indicate that performance is compromised at both extremes, with the optimal outcome attained when the max distance parameter is set to 1.0.

\begin{table*}[t]
\centering
\begin{minipage}{0.45\linewidth}
\centering
\caption{Debias Strategy. P. Acc stands for Pairwise Accuracy}
\label{tab:debias}
\begin{tabular}{lccc}
\hline
\textbf{Group By} & \textbf{AUC} & \textbf{P. Acc} & $\boldsymbol{\rho}$ \\
\hline
None          & 0.739 & 0.630 & 0.400 \\
Pub. Time  & 0.736 & 0.638 & 0.392 \\
Keyword  & 0.556 & 0.620 & 0.007 \\
Hier. Cluster & 0.753 & 0.639 & 0.401 \\
\textbf{Time+Hier.} & \textbf{0.782} & \textbf{0.649} & \textbf{0.432} \\
\hline
\end{tabular}
\end{minipage}
\hfill
\begin{minipage}{0.48\linewidth}
\centering
\caption{Effectiveness of RTS as a Quality Signal. P. Acc stands for Pairwise Accuracy}
\label{tab:RTS}
\begin{tabular}{lccc}
\hline
\textbf{Signal} & \textbf{AUC} & \textbf{P. Acc} & $\boldsymbol{\rho}$ \\
\hline
Mean     & 0.753 & 0.586 & 0.385 \\
Weighted   & 0.754 & 0.598 & 0.402 \\
Median   & 0.757 & 0.599 & 0.398 \\
Mode     & 0.730 & 0.564 & 0.343 \\
\textbf{RTS (ours)} & \textbf{0.782} & \textbf{0.609} & \textbf{0.432} \\
\hline
\end{tabular}
\end{minipage}
\end{table*}

\subsection{Benefit of the proposed RTS}

Table~\ref{tab:RTS} compares RTS with common aggregation strategies. Simple averaging (Mean) and its weighted variant show only marginal differences, indicating that naïve incorporation of confidence contributes little. Median aggregation performs slightly better, but the gain remains limited. Mode aggregation, on the other hand, leads to the weakest performance, as it discards the distributional nuances of reviewer scores. In comparison, RTS achieves consistently stronger results. Its probabilistic treatment of reviewer confidence allows more informative integration of scores, resulting in a more discriminative and reliable quality signal than all traditional baselines.

\subsection{Learning with Difficulty-Aware Supervision}
\label{sec:learn_difficulty}
\begin{wraptable}[8]{r}{0.45\textwidth}  
\centering
\captionof{table}{Impact of curriculum learning (``w/o'' denotes without, ``w/'' denotes with).}  
\label{tab:cur_learning}
\begin{tabular}{lcc}
\hline
\textbf{Setting} & \textbf{AUC} & \boldsymbol{$\rho$}  \\
\hline
w/o Curriculum & 0.770 & 0.418  \\
w/  Curriculum & 0.782 & 0.432 \\
\hline
\end{tabular}
\end{wraptable}

To study the effect of pairwise difficulty, we group training pairs into buckets by the $RTS$ gap $\Delta_{ab}$ (detailed in Appendix~\ref{app:bucket_dist}), where larger gaps indicate easier pairs and smaller gaps correspond to harder ones. Figure~\ref{fig:curriculum} shows that performance consistently improves as the proportion of easy pairs increases. This finding suggests that simple comparisons provide more stable and reliable learning signals, whereas emphasizing hard pairs tends to introduce noise and degrade performance. Building on this observation, we further adopt a curriculum learning strategy. In the early training stages, sampling is biased toward easy pairs, allowing the model to first capture coarse distinctions. Over time, harder pairs are gradually introduced, enabling the model to refine its judgment on subtle quality differences. The results in Tab.~\ref{tab:cur_learning} verify that curriculum learning provides measurable improvements, supporting it as a useful complement to difficulty-aware supervision.


\section{From ICLR to NeurIPS: A Generalization Case Study}

To further examine the generalization ability of our model, we conduct a cross-venue evaluation on NeurIPS. Following the same protocol as for NAIDv2, we construct a NeurIPS test set comprising 13,223 samples, covering submissions from 2021 to 2024. Unlike ICLR, NeurIPS exhibits an imbalance in review availability: reviews of accepted papers always appear in public, whereas reviews of rejected papers emerge only when authors choose to disclose them. This practice inflates the proportion of accepted papers in the accessible record relative to the true acceptance rate of about 20\%, thereby producing a strongly skewed distribution. This imbalance renders previous tasks on the NeurIPS dataset less meaningful.

\begin{figure}[h]
  \centering
    \includegraphics[width=0.95\linewidth]{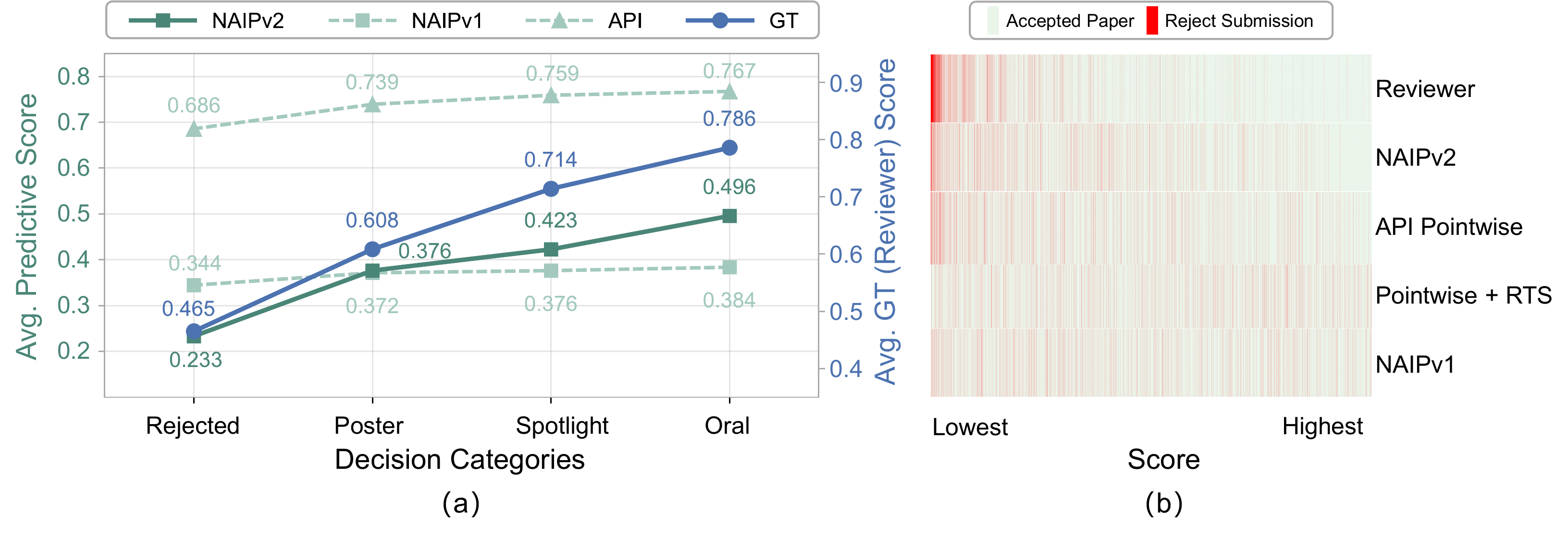}
  \caption{Cross-venue evaluation on NeurIPS. (a) Predicted scores follow the hierarchical structure of conference decisions. (b) Distribution of rejected papers across score ranges.}
  \label{fig:cross-venue}
\end{figure}

Considering that NeurIPS organizes submissions into a hierarchy, with rejected papers at the lowest tier and accepted papers further divided into poster, spotlight, and oral presentations of increasing contribution and quality, it is natural to expect stronger papers to receive higher scores and be placed into more prestigious categories. We examine whether our predicted evaluation scores (sigmoid normalized) align with this hierarchy, and as shown in Fig.~\ref{fig:cross-venue} (a), the model outputs follow a clear upward trend from rejected to poster, spotlight, and oral papers compared to NAIPv1 and the pointwise API-based method, demonstrating strong consistency with the actual conference decisions.

In Fig.~\ref{fig:cross-venue} (b), we show the distribution of rejected papers across score ranges. Each row corresponds to the min-to-max assigned scores. Light green denotes accepted papers, while deep red denotes rejections. The reviewer row can be regarded as the ground-truth distribution. As expected, most rejections cluster in the lower-score region, yet some appear in higher ranges, reflecting the valuable role of expert preferences in calibrating the review process. NAIPv2 exhibits a trend similar to reviewer assessments, with lower-score regions containing more rejected papers, while the high-score regions contain fewer rejections than the reviewer-assigned scores. The API-based approach shows a weaker tendency, yet still outperforms stochastic prediction. Owing to inconsistencies, NAIPv1 and direct RTS regression perform poorly, as rejected papers are distributed almost randomly across score ranges. This case study shows that NAIPv2 achieves stable generalization and consistent alignment with conference outcomes, suggesting promising directions for broader applications.

\section{Conclusion}
This work presents NAIPv2, a debiased and scalable framework for automated paper quality estimation. By employing pairwise learning within domain-year groups and introducing a confidence-aware probabilistic signal, NAIPv2 mitigates score inconsistencies while enabling efficient linear-time inference. On the curated ICLR test set, the model achieves state-of-the-art performance, demonstrating both accuracy and efficiency compared to prior approaches. Moreover, in a cross-venue evaluation on NeurIPS, NAIPv2 predicts a clear monotonic trend from rejected to oral papers and accurately places less rejections in the high-score regions, demonstrating consistency with human judgments and robustness under distributional shifts. Taken together, these results highlight NAIPv2 as an exploratory step toward robust scalable scientific intelligence systems, while leaving open important avenues for future research on quality prediction and broader academic applications.

\subsubsection*{Acknowledgments}
This research was supported by the Fund of the National Natural Science Foundation of China (Grant No.62576177, 62206134), the Fundamental Research Funds for the Central Universities 070-63233084, and the Tianjin Key Laboratory of Visual Computing and Intelligent Perception (VCIP). Computation is supported by the Supercomputing Center of Nankai University (NKSC). This work was supported by the National Science Fund of China under Grant No. 62361166670.

\bibliography{iclr2026_conference}

\begin{thebibliography}{49}
\providecommand{\natexlab}[1]{#1}
\providecommand{\url}[1]{\texttt{#1}}
\expandafter\ifx\csname urlstyle\endcsname\relax
  \providecommand{\doi}[1]{doi: #1}\else
  \providecommand{\doi}{doi: \begingroup \urlstyle{rm}\Url}\fi

\bibitem[Ali et~al.(2025)Ali, Jat~Baloch, Naveed, Nigar, Almalki, Rasool, Gedfew, and Arafat]{ali2025advanced}
Awais Ali, Muhammad~Yousuf Jat~Baloch, Muhammad Naveed, Anam Nigar, Abdulrahman~Seraj Almalki, Ayesha~Ghulam Rasool, Meseret~Abeje Gedfew, and Ahmed~A Arafat.
\newblock Advanced satellite-based remote sensing and data analytics for precision water resource management and agricultural optimization.
\newblock \emph{Scientific Reports}, 15\penalty0 (1):\penalty0 27527, 2025.

\bibitem[Baek et~al.(2024)Baek, Jauhar, Cucerzan, and Hwang]{baek2024researchagent}
Jinheon Baek, Sujay~Kumar Jauhar, Silviu Cucerzan, and Sung~Ju Hwang.
\newblock Researchagent: Iterative research idea generation over scientific literature with large language models.
\newblock \emph{arXiv preprint arXiv:2404.07738}, 2024.

\bibitem[Bar-Joseph et~al.(2001)Bar-Joseph, Gifford, and Jaakkola]{bar2001fast}
Ziv Bar-Joseph, David~K Gifford, and Tommi~S Jaakkola.
\newblock Fast optimal leaf ordering for hierarchical clustering.
\newblock \emph{Bioinformatics}, 17\penalty0 (suppl\_1):\penalty0 S22--S29, 2001.

\bibitem[Bharti et~al.(2021)Bharti, Ranjan, Ghosal, Agrawal, and Ekbal]{bharti2021peerassist}
Prabhat~Kumar Bharti, Shashi Ranjan, Tirthankar Ghosal, Mayank Agrawal, and Asif Ekbal.
\newblock Peerassist: leveraging on paper-review interactions to predict peer review decisions.
\newblock In \emph{International Conference on Asian Digital Libraries}, pp.\  421--435. Springer, 2021.

\bibitem[Cao et~al.(2007)Cao, Qin, Liu, Tsai, and Li]{cao2007learning}
Zhe Cao, Tao Qin, Tie-Yan Liu, Ming-Feng Tsai, and Hang Li.
\newblock Learning to rank: from pairwise approach to listwise approach.
\newblock In \emph{Proceedings of the 24th international conference on Machine learning}, pp.\  129--136, 2007.

\bibitem[Chen et~al.(2024{\natexlab{a}})Chen, Xiao, Zhang, Luo, Lian, and Liu]{chen2024bge}
Jianlv Chen, Shitao Xiao, Peitian Zhang, Kun Luo, Defu Lian, and Zheng Liu.
\newblock Bge m3-embedding: Multi-lingual, multi-functionality, multi-granularity text embeddings through self-knowledge distillation.
\newblock \emph{arXiv preprint arXiv:2402.03216}, 2024{\natexlab{a}}.

\bibitem[Chen et~al.(2024{\natexlab{b}})Chen, Yang, Chen, Chen, Liang, and Li]{chen2024revisiting}
Zhenyuan Chen, Lingfeng Yang, Shuo Chen, Zhaowei Chen, Jiajun Liang, and Xiang Li.
\newblock Revisiting prompt pretraining of vision-language models.
\newblock \emph{arXiv preprint arXiv:2409.06166}, 2024{\natexlab{b}}.

\bibitem[Cortes \& Lawrence(2021)Cortes and Lawrence]{cortes2021inconsistency}
Corinna Cortes and Neil~D Lawrence.
\newblock Inconsistency in conference peer review: Revisiting the 2014 neurips experiment.
\newblock \emph{arXiv preprint arXiv:2109.09774}, 2021.

\bibitem[D'Arcy et~al.(2024)D'Arcy, Hope, Birnbaum, and Downey]{d2024marg}
Mike D'Arcy, Tom Hope, Larry Birnbaum, and Doug Downey.
\newblock Marg: Multi-agent review generation for scientific papers.
\newblock \emph{arXiv preprint arXiv:2401.04259}, 2024.

\bibitem[de~Winter(2024)]{de2024can}
Joost de~Winter.
\newblock Can chatgpt be used to predict citation counts, readership, and social media interaction? an exploration among 2222 scientific abstracts.
\newblock \emph{Scientometrics}, 129\penalty0 (4):\penalty0 2469--2487, 2024.

\bibitem[Dougherty \& Horne(2022)Dougherty and Horne]{dougherty2022citation}
Michael~R Dougherty and Zachary Horne.
\newblock Citation counts and journal impact factors do not capture some indicators of research quality in the behavioural and brain sciences.
\newblock \emph{Royal Society Open Science}, 9\penalty0 (8):\penalty0 220334, 2022.

\bibitem[Dubey et~al.(2024)Dubey, Jauhri, Pandey, Kadian, Al-Dahle, Letman, Mathur, Schelten, Yang, Fan, et~al.]{dubey2024llama}
Abhimanyu Dubey, Abhinav Jauhri, Abhinav Pandey, Abhishek Kadian, Ahmad Al-Dahle, Aiesha Letman, Akhil Mathur, Alan Schelten, Amy Yang, Angela Fan, et~al.
\newblock The llama 3 herd of models.
\newblock \emph{arXiv e-prints}, pp.\  arXiv--2407, 2024.

\bibitem[F1000Research et~al.(2012)F1000Research, Library, Ireland, and of~Calgary]{american2012san}
F1000Research, Iowa State~University Library, Research Ireland, and University of~Calgary.
\newblock San francisco declaration on research assessment (dora).
\newblock 2012.

\bibitem[Ghosal et~al.(2019)Ghosal, Verma, Ekbal, and Bhattacharyya]{ghosal2019deepsentipeer}
Tirthankar Ghosal, Rajeev Verma, Asif Ekbal, and Pushpak Bhattacharyya.
\newblock Deepsentipeer: Harnessing sentiment in review texts to recommend peer review decisions.
\newblock In \emph{Proceedings of the 57th Annual Meeting of the Association for Computational Linguistics}, pp.\  1120--1130, 2019.

\bibitem[Girshick(2015)]{girshick2015fast}
Ross Girshick.
\newblock Fast r-cnn.
\newblock In \emph{Proceedings of the IEEE international conference on computer vision}, pp.\  1440--1448, 2015.

\bibitem[Hicks et~al.(2015)Hicks, Wouters, Waltman, De~Rijcke, and Rafols]{hicks2015bibliometrics}
Diana Hicks, Paul Wouters, Ludo Waltman, Sarah De~Rijcke, and Ismael Rafols.
\newblock Bibliometrics: the leiden manifesto for research metrics.
\newblock \emph{Nature}, 520\penalty0 (7548):\penalty0 429--431, 2015.

\bibitem[Hinton et~al.(2015)Hinton, Vinyals, and Dean]{hinton2015distilling}
Geoffrey Hinton, Oriol Vinyals, and Jeff Dean.
\newblock Distilling the knowledge in a neural network.
\newblock \emph{arXiv preprint arXiv:1503.02531}, 2015.

\bibitem[H{\"o}pner et~al.(2025)H{\"o}pner, Eshuijs, Alivanistos, Zamprogno, and Tiddi]{hopner2025automatic}
Niklas H{\"o}pner, Leon Eshuijs, Dimitrios Alivanistos, Giacomo Zamprogno, and Ilaria Tiddi.
\newblock Automatic evaluation metrics for artificially generated scientific research.
\newblock \emph{arXiv preprint arXiv:2503.05712}, 2025.

\bibitem[Hu et~al.(2022)Hu, Shen, Wallis, Allen-Zhu, Li, Wang, Wang, Chen, et~al.]{hu2022lora}
Edward~J Hu, Yelong Shen, Phillip Wallis, Zeyuan Allen-Zhu, Yuanzhi Li, Shean Wang, Lu~Wang, Weizhu Chen, et~al.
\newblock Lora: Low-rank adaptation of large language models.
\newblock \emph{ICLR}, 1\penalty0 (2):\penalty0 3, 2022.

\bibitem[Idahl \& Ahmadi(2024)Idahl and Ahmadi]{idahl2024openreviewer}
Maximilian Idahl and Zahra Ahmadi.
\newblock Openreviewer: A specialized large language model for generating critical scientific paper reviews.
\newblock \emph{arXiv preprint arXiv:2412.11948}, 2024.

\bibitem[Kang et~al.(2018)Kang, Ammar, Dalvi, Van~Zuylen, Kohlmeier, Hovy, and Schwartz]{kang2018dataset}
Dongyeop Kang, Waleed Ammar, Bhavana Dalvi, Madeleine Van~Zuylen, Sebastian Kohlmeier, Eduard Hovy, and Roy Schwartz.
\newblock A dataset of peer reviews (peerread): Collection, insights and nlp applications.
\newblock \emph{arXiv preprint arXiv:1804.09635}, 2018.

\bibitem[Kuckreja et~al.(2024)Kuckreja, Danish, Naseer, Das, Khan, and Khan]{kuckreja2024geochat}
Kartik Kuckreja, Muhammad~Sohail Danish, Muzammal Naseer, Abhijit Das, Salman Khan, and Fahad~Shahbaz Khan.
\newblock Geochat: Grounded large vision-language model for remote sensing.
\newblock In \emph{Proceedings of the IEEE/CVF Conference on Computer Vision and Pattern Recognition}, pp.\  27831--27840, 2024.

\bibitem[Li et~al.(2024{\natexlab{a}})Li, Li, Li, Zhang, Dai, Hou, Cheng, and Yang]{li2024sm3det}
Yuxuan Li, Xiang Li, Yunheng Li, Yicheng Zhang, Yimian Dai, Qibin Hou, Ming-Ming Cheng, and Jian Yang.
\newblock Sm3det: A unified model for multi-modal remote sensing object detection.
\newblock \emph{arXiv preprint arXiv:2412.20665}, 2024{\natexlab{a}}.

\bibitem[Li et~al.(2025)Li, Zhang, Tang, Dai, Cheng, Li, and Yang]{li2025vitp}
Yuxuan Li, Yicheng Zhang, Wenhao Tang, Yimian Dai, Ming-Ming Cheng, Xiang Li, and Jian Yang.
\newblock Visual instruction pretraining for domain-specific foundation models.
\newblock \emph{arXiv preprint arXiv:2509.17562}, 2025.

\bibitem[Li et~al.(2023)Li, Li, Yang, Zhao, Song, Luo, Li, and Yang]{li2023curriculum}
Zheng Li, Xiang Li, Lingfeng Yang, Borui Zhao, Renjie Song, Lei Luo, Jun Li, and Jian Yang.
\newblock Curriculum temperature for knowledge distillation.
\newblock In \emph{Proceedings of the AAAI Conference on Artificial Intelligence}, volume~37, pp.\  1504--1512, 2023.

\bibitem[Li et~al.(2024{\natexlab{b}})Li, Li, Fu, Zhang, Wang, Chen, and Yang]{li2024promptkd}
Zheng Li, Xiang Li, Xinyi Fu, Xin Zhang, Weiqiang Wang, Shuo Chen, and Jian Yang.
\newblock Promptkd: Unsupervised prompt distillation for vision-language models.
\newblock In \emph{Proceedings of the IEEE/CVF Conference on Computer Vision and Pattern Recognition}, pp.\  26617--26626, 2024{\natexlab{b}}.

\bibitem[Li et~al.(2024{\natexlab{c}})Li, Song, Cheng, Li, and Yang]{li2024advancing}
Zheng Li, Yibing Song, Ming-Ming Cheng, Xiang Li, and Jian Yang.
\newblock Advancing textual prompt learning with anchored attributes.
\newblock \emph{arXiv preprint arXiv:2412.09442}, 1, 2024{\natexlab{c}}.

\bibitem[Liu et~al.(2009)]{liu2009learning}
Tie-Yan Liu et~al.
\newblock Learning to rank for information retrieval.
\newblock \emph{Foundations and Trends{\textregistered} in Information Retrieval}, 3\penalty0 (3):\penalty0 225--331, 2009.

\bibitem[Lu et~al.(2024)Lu, Lu, Lange, Foerster, Clune, and Ha]{lu2024ai}
Chris Lu, Cong Lu, Robert~Tjarko Lange, Jakob Foerster, Jeff Clune, and David Ha.
\newblock The ai scientist: Towards fully automated open-ended scientific discovery.
\newblock \emph{arXiv preprint arXiv:2408.06292}, 2024.

\bibitem[Maslej et~al.(2025)Maslej, Fattorini, Perrault, Gil, Parli, Kariuki, Capstick, Reuel, Brynjolfsson, Etchemendy, et~al.]{maslej2025artificial}
Nestor Maslej, Loredana Fattorini, Raymond Perrault, Yolanda Gil, Vanessa Parli, Njenga Kariuki, Emily Capstick, Anka Reuel, Erik Brynjolfsson, John Etchemendy, et~al.
\newblock Artificial intelligence index report 2025.
\newblock \emph{arXiv preprint arXiv:2504.07139}, 2025.

\bibitem[McInnes et~al.(2017)McInnes, Healy, Astels, et~al.]{mcinnes2017hdbscan}
Leland McInnes, John Healy, Steve Astels, et~al.
\newblock hdbscan: Hierarchical density based clustering.
\newblock \emph{J. Open Source Softw.}, 2\penalty0 (11):\penalty0 205, 2017.

\bibitem[M{\"u}llner(2011)]{mullner2011modern}
Daniel M{\"u}llner.
\newblock Modern hierarchical, agglomerative clustering algorithms.
\newblock \emph{arXiv preprint arXiv:1109.2378}, 2011.

\bibitem[Si et~al.(2024)Si, Yang, and Hashimoto]{si2024can}
Chenglei Si, Diyi Yang, and Tatsunori Hashimoto.
\newblock Can llms generate novel research ideas? a large-scale human study with 100+ nlp researchers.
\newblock \emph{arXiv preprint arXiv:2409.04109}, 2024.

\bibitem[Staudinger et~al.(2024)Staudinger, Kusa, Piroi, and Hanbury]{staudinger2024analysis}
Moritz Staudinger, Wojciech Kusa, Florina Piroi, and Allan Hanbury.
\newblock An analysis of tasks and datasets in peer reviewing.
\newblock In \emph{Proceedings of the Fourth Workshop on Scholarly Document Processing (SDP 2024)}, pp.\  257--268, 2024.

\bibitem[Thelwall(2025)]{thelwall2025research}
Mike Thelwall.
\newblock Research quality evaluation by ai in the era of large language models: advantages, disadvantages, and systemic effects--an opinion paper.
\newblock \emph{Scientometrics}, pp.\  1--13, 2025.

\bibitem[Wang et~al.(2024)Wang, Guo, Yao, Zhang, Zhang, Wu, Zhang, Dai, Wen, Ye, et~al.]{wang2024autosurvey}
Yidong Wang, Qi~Guo, Wenjin Yao, Hongbo Zhang, Xin Zhang, Zhen Wu, Meishan Zhang, Xinyu Dai, Qingsong Wen, Wei Ye, et~al.
\newblock Autosurvey: Large language models can automatically write surveys.
\newblock \emph{Advances in neural information processing systems}, 37:\penalty0 115119--115145, 2024.

\bibitem[Weng et~al.(2025)Weng, Zhu, Bao, Zhang, Wang, Zhang, and Yang]{weng2025cycleresearcher}
Yixuan Weng, Minjun Zhu, Guangsheng Bao, Hongbo Zhang, Jindong Wang, Yue Zhang, and Linyi Yang.
\newblock Cycleresearcher: Improving automated research via automated review.
\newblock In \emph{The Thirteenth International Conference on Learning Representations}, 2025.
\newblock URL \url{https://openreview.net/forum?id=bjcsVLoHYs}.

\bibitem[Wu et~al.(2024)Wu, Zhang, Li, Chen, Liang, Yang, and Li]{wu2024cascade}
Ge~Wu, Xin Zhang, Zheng Li, Zhaowei Chen, Jiajun Liang, Jian Yang, and Xiang Li.
\newblock Cascade prompt learning for vision-language model adaptation.
\newblock In \emph{European Conference on Computer Vision}, pp.\  304--321. Springer, 2024.

\bibitem[Xia et~al.(2023)Xia, Li, and Li]{xia2023review}
Wanjun Xia, Tianrui Li, and Chongshou Li.
\newblock A review of scientific impact prediction: tasks, features and methods.
\newblock \emph{Scientometrics}, 128\penalty0 (1):\penalty0 543--585, 2023.

\bibitem[Yu et~al.(2024)Yu, Ding, Tan, Luo, Weng, Gong, Zeng, Cui, Han, Sun, et~al.]{yu2024automated}
Jianxiang Yu, Zichen Ding, Jiaqi Tan, Kangyang Luo, Zhenmin Weng, Chenghua Gong, Long Zeng, Renjing Cui, Chengcheng Han, Qiushi Sun, et~al.
\newblock Automated peer reviewing in paper sea: Standardization, evaluation, and analysis.
\newblock \emph{arXiv preprint arXiv:2407.12857}, 2024.

\bibitem[Zhang et~al.(2025{\natexlab{a}})Zhang, Bao, Du, Zhao, Zhang, Bao, and Yang]{zhang2025re}
Daoze Zhang, Zhijian Bao, Sihang Du, Zhiyi Zhao, Kuangling Zhang, Dezheng Bao, and Yang Yang.
\newblock Re2: A consistency-ensured dataset for full-stage peer review and multi-turn rebuttal discussions.
\newblock \emph{arXiv preprint arXiv:2505.07920}, 2025{\natexlab{a}}.

\bibitem[Zhang et~al.(2025{\natexlab{b}})Zhang, Yang, Li, Yang, Cheng, and Li]{zhang2025rsar}
Xin Zhang, Xue Yang, Yuxuan Li, Jian Yang, Ming-Ming Cheng, and Xiang Li.
\newblock Rsar: Restricted state angle resolver and rotated sar benchmark.
\newblock In \emph{Proceedings of the Computer Vision and Pattern Recognition Conference}, pp.\  7416--7426, 2025{\natexlab{b}}.

\bibitem[Zhang et~al.(2025{\natexlab{c}})Zhang, Li, Long, Zhang, Lin, Yang, Xie, Yang, Liu, Lin, et~al.]{zhang2025qwen3}
Yanzhao Zhang, Mingxin Li, Dingkun Long, Xin Zhang, Huan Lin, Baosong Yang, Pengjun Xie, An~Yang, Dayiheng Liu, Junyang Lin, et~al.
\newblock Qwen3 embedding: Advancing text embedding and reranking through foundation models.
\newblock \emph{arXiv preprint arXiv:2506.05176}, 2025{\natexlab{c}}.

\bibitem[Zhang et~al.(2025{\natexlab{d}})Zhang, Zhang, Ji, Hua, Haber, Cao, and Liang]{zhang2025replication}
Yaohui Zhang, Haijing Zhang, Wenlong Ji, Tianyu Hua, Nick Haber, Hancheng Cao, and Weixin Liang.
\newblock From replication to redesign: Exploring pairwise comparisons for llm-based peer review.
\newblock \emph{arXiv preprint arXiv:2506.11343}, 2025{\natexlab{d}}.

\bibitem[Zhao et~al.(2024)Zhao, Zhang, Cao, Cheng, Yang, and Li]{zhao2024literature}
Penghai Zhao, Xin Zhang, Jiayue Cao, Ming-Ming Cheng, Jian Yang, and Xiang Li.
\newblock A literature review of literature reviews in pattern analysis and machine intelligence.
\newblock \emph{arXiv preprint arXiv:2402.12928}, 2024.

\bibitem[Zhao et~al.(2025)Zhao, Xing, Dou, Tian, Tai, Yang, Cheng, and Li]{zhao2025words}
Penghai Zhao, Qinghua Xing, Kairan Dou, Jinyu Tian, Ying Tai, Jian Yang, Ming-Ming Cheng, and Xiang Li.
\newblock From words to worth: Newborn article impact prediction with llm.
\newblock In \emph{Proceedings of the AAAI Conference on Artificial Intelligence}, volume~39, pp.\  1183--1191, 2025.

\bibitem[Zhao \& Feng(2022)Zhao and Feng]{zhao2022utilizing}
Qihang Zhao and Xiaodong Feng.
\newblock Utilizing citation network structure to predict paper citation counts: A deep learning approach.
\newblock \emph{Journal of Informetrics}, 16\penalty0 (1):\penalty0 101235, 2022.

\bibitem[Zhou et~al.(2024)Zhou, Chen, and Yu]{zhou2024llm}
Ruiyang Zhou, Lu~Chen, and Kai Yu.
\newblock Is llm a reliable reviewer? a comprehensive evaluation of llm on automatic paper reviewing tasks.
\newblock In \emph{Proceedings of the 2024 joint international conference on computational linguistics, language resources and evaluation (LREC-COLING 2024)}, pp.\  9340--9351, 2024.

\bibitem[Zhu et~al.(2025)Zhu, Weng, Yang, and Zhang]{zhu-2025-deepreview}
Minjun Zhu, Yixuan Weng, Linyi Yang, and Yue Zhang.
\newblock Deepreview: Improving llm-based paper review with human-like deep thinking process.
\newblock In \emph{Proceedings of the 63rd Annual Meeting of the Association for Computational Linguistics (Volume 1: Long Papers)}, pp.\  29330--29355, Vienna, Austria, July 2025. Association for Computational Linguistics.
\newblock ISBN 979-8-89176-251-0.
\newblock \doi{10.18653/v1/2025.acl-long.1420}.
\newblock URL \url{https://aclanthology.org/2025.acl-long.1420/}.

\end{thebibliography}
\bibliographystyle{iclr2026_conference}

\newpage
\appendix

\section{Ethical Statement}
The scores presented in this work are generated by machine learning models trained on historical peer-review data and should be understood as black-box predictions, not as objective or authoritative measures of scientific quality. These scores are not intended, nor should they be used, to replace human peer review or editorial decision-making. Our contribution is solely for research purposes, with the aim of studying peer-review dynamics and developing methods for literature intelligence systems. The authors bear no responsibility for any use of the predicted scores in real-world decision processes.

\section{Reproducibility Statement}
We fix the random seed to $42$ that controls all randomness in \texttt{random}, \texttt{NumPy}, and \texttt{PyTorch}, including CUDA backends. 
This minimizes nondeterminism during training and evaluation.

\section{Disclosure of LLM Usage}
ChatGPT was employed during the experimental stage as an auxiliary tool for code debugging, code development, and data visualization. All code was thoroughly reviewed by the authors to ensure accuracy and reliability. ChatGPT also assisted in polishing the language of the manuscript, but was not involved in substantive writing. The authors take full responsibility for any issues arising from the use of ChatGPT.

\section{Debiased Pairwise Learning Objectives}
\label{app:loss}
In this section we broaden our investigation beyond the main loss and systematically examine 
alternative objectives for pairwise learning. The motivation is to understand the design space 
of loss functions and the trade-offs they entail, rather than to commit to a single formulation. 
Formally, given a pair $(a,b)$ with binary label $z_{ab}=\mathbb{I}[{RTS}a>{RTS}b]$, outputs 
$y_a, y_b$, and difference $d_{ab}=y_a-y_b$, we consider the following representative families.  

\paragraph{(A) Likelihood-based objectives.}  
These methods interpret the pairwise preference as a probability that $a$ is better than $b$, 
and optimize a proper scoring rule between predicted and observed preference.

\emph{Thurstone–Probit.}  
Replacing the sigmoid with a probit link yields  
\begin{equation}
\hat{z}_{ab}^{\textsf{probit}} = \Phi\!\left(\tfrac{d_{ab}}{\tau}\right), \quad
\mathcal{L}_{\textsf{Probit}}(a,b) =
- \Big(z_{ab}\log \hat{z}_{ab}^{\textsf{probit}} + (1-z_{ab})\log(1-\hat{z}_{ab}^{\textsf{probit}})\Big),
\label{eq:probit-loss}
\end{equation}
where $\Phi(\cdot)$ is the standard normal CDF and $\tau$ controls the slope.  
This treats preference as arising from a latent Gaussian noise model, offering smoother gradients when comparisons are uncertain.  

\emph{Pairwise Brier.}  
Instead of a log loss, we minimize squared error between predicted probability and label:  
\begin{equation}
\mathcal{L}_{\textsf{Brier}}(a,b) = \big(\sigma(d_{ab})-z_{ab}\big)^2.
\label{eq:brier-loss}
\end{equation}
This penalizes deviations quadratically and is less harsh than cross-entropy when predictions are confident but wrong.  

\paragraph{(B) Margin- and regression-style objectives.}  
These objectives work directly on the score difference $d_{ab}$, either enforcing a margin 
or softly aligning it with the review gap $\Delta_{ab}$.  

\emph{Hinge margin.}  
With labels $y_{ab}=2z_{ab}-1 \in \{-1,+1\}$, we require $d_{ab}$ to exceed a margin $m$:  
\begin{equation}
\mathcal{L}_{\textsf{Hinge}}(a,b) = \big[m - y_{ab}\,d_{ab}\big]_+, 
\quad [x]_+ = \max(x,0).
\label{eq:hinge-loss}
\end{equation}
This loss is zero if the preferred item is already ahead by at least $m$, 
and otherwise grows linearly—encouraging confident separation.  

\emph{Huber-on-difference.}  
Here we compare the predicted difference to the observed review score gap $\Delta_{ab}$.  
Residual $r_{ab} = d_{ab}-\kappa\,\Delta_{ab}$ is penalized with a robust Huber function:  
\begin{equation}
\mathcal{L}_{\textsf{HuberDiff}}(a,b) =
\begin{cases}
\tfrac{1}{2}r_{ab}^2, & |r_{ab}|\le\delta, \\[3pt]
\delta\!\left(|r_{ab}|-\tfrac{\delta}{2}\right), & |r_{ab}|>\delta.
\end{cases}
\label{eq:huber-loss}
\end{equation}
This encourages alignment with the magnitude of human score gaps, 
while limiting the influence of outliers beyond threshold $\delta$.  

\paragraph{(C) Hybrid objectives.}  
These combine a directional likelihood with a light regression regularizer, aiming to 
preserve ordering while stabilizing score magnitudes.  

\emph{Rank-then-Regress.}  
We augment RankNet with an $L_2$ penalty on the difference–gap residual:  
\begin{equation}
\mathcal{L}_{\textsf{RtR}}(a,b) = 
- \Big(z_{ab}\log\sigma(d_{ab}) + (1-z_{ab})\log(1-\sigma(d_{ab}))\Big) 
+ \lambda\,(d_{ab}-\kappa\,\Delta_{ab})^2,
\label{eq:rtr-loss}
\end{equation}
where $\lambda$ controls the regression weight.  
This formulation preserves pairwise preference consistency while encouraging smoother calibration within domain–year subgroups.

\begin{table}[t]
\centering
\caption{Comparison of different loss functions for pairwise learning.}
\label{tab:loss_compare}
\begin{tabular}{llccccc}
\hline
\textbf{Category} & \textbf{Loss} & \textbf{Acc} & \textbf{F1} & \textbf{AUC} & \textbf{NDCG} & $\boldsymbol{\rho}$ \\
\hline
Likelihood-based   & BCE (\textit{adopted})  & 0.706 & \textbf{0.609} & \textbf{0.782} & 0.771 &\textbf{0.432} \\
                   & Thurstone--Probit   & 0.689 & 0.594 & 0.756 & 0.762 & 0.393 \\
                   & Pairwise Brier      & 0.606 & 0.541 & 0.685 & 0.750 & 0.293 \\
\hline
Margin / Regression & Hinge Margin        & \textbf{0.711} & 0.606 & 0.772 & \textbf{0.783} & 0.420 \\
                   & Huber-on-difference & 0.531 & 0.548 & 0.699 & 0.758 & 0.288 \\
\hline
Hybrid             & Rank-then-Regress   & 0.681 & 0.591 & 0.765 & 0.789 & 0.396 \\
\hline
\end{tabular}
\end{table}

Overall, results in Tab.~\ref{tab:loss_compare} support our choice of a purely pairwise likelihood objective as the most stable formulation. Importantly, once regression-style supervision is introduced, performance degrades across metrics, suggesting that absolute review scores suffer from scale inconsistencies across domains and years. This reinforces the view that enforcing numerical alignment is detrimental, whereas ordinal-only objectives provide more robust and transferable signals.

\section{Data Efficiency Analysis}
\label{app:data_efficiency}

\begin{figure}[ht]
\centering
\includegraphics[width=1.0\textwidth]{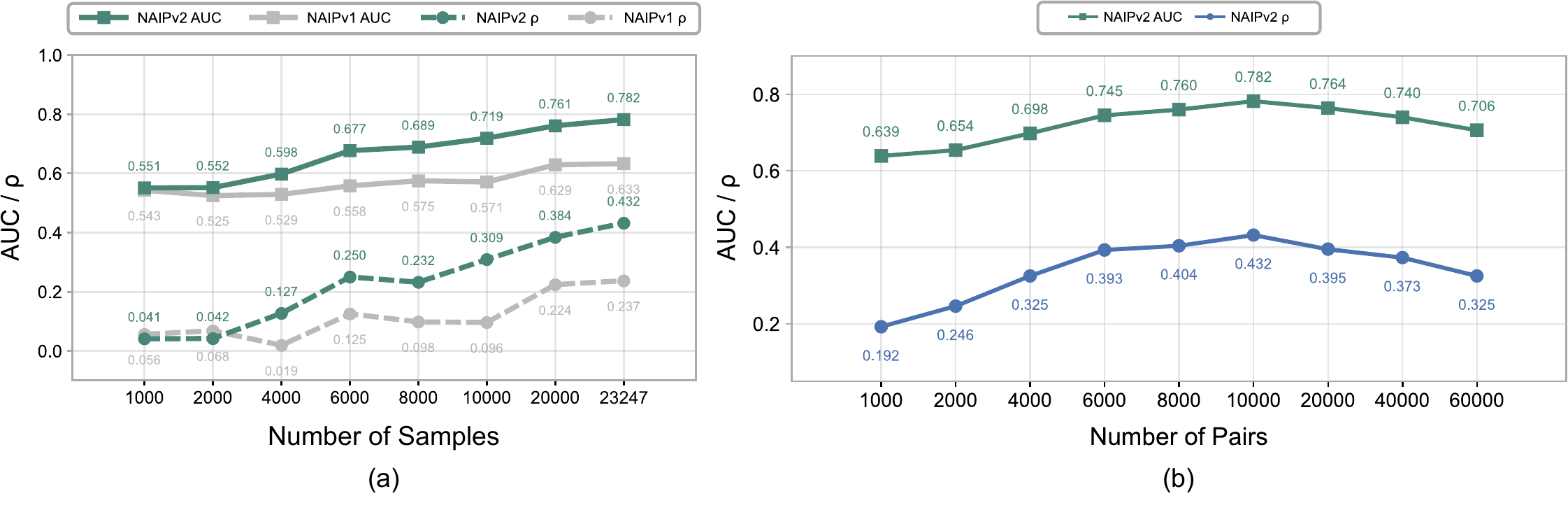}
\caption{Effect of data size on model performance. 
(a) Performance comparison of NAIPv1 and NAIPv2 with different numbers of samples, measured by AUC and Spearman’s $\rho$. 
(b) Performance of NAIPv2 with different numbers of training pairs.}
\label{fig:data_efficiency}
\end{figure}

We analyze the effect of data size on NAIPv2 performance from two perspectives.  
Figure~\ref{fig:data_efficiency}(a) compares NAIPv1 and NAIPv2 under varying sample sizes. Both models benefit from more data, and NAIPv2 consistently outperforms NAIPv1. However, due to the scarcity of reviewed data, we could not scale sample size further. Figure~\ref{fig:data_efficiency}(b) examines NAIPv2 trained with different numbers of pairs. The model shows clear improvements as the number of pairs increases up to about 10,000. Beyond this point, performance saturates, and additional pairs do not provide significant gains; instead, they may even degrade performance due to overfitting when the task difficulty becomes too low. These results suggest that while more data is generally beneficial, NAIPv2 requires only about 10,000 pairs to reach optimal performance, and further increasing the pair data size is unnecessary and potentially harmful.

\section{Normalization in the Computation of RTS}
\label{app:RTS_norm}

In computing the RTS, we do not directly rely on the raw review scores; instead, both the scores and the confidence levels are normalized. The normalization of scores is straightforward and follows a standard min–max scaling scheme:
\begin{equation}
\tilde{s}_i \;=\; \frac{s_i - \min_j s_j}{\max_j s_j - \min_j s_j},
\end{equation}
where $s_i$ denotes the raw score given by reviewer $i$, and $\tilde{s}_i \in [0,1]$ is the normalized score.

For confidence levels, however, the choice of normalization strategy is less obvious. A central question is how the lowest reported confidence should be mapped on the normalized scale—e.g., whether it should correspond to 0.2, 0.5, 0.8, or other values. To allow for this flexibility, we define the normalization as:
\begin{equation}
\tilde{c}_i \;=\; \alpha \;+\; (1-\alpha)\,\frac{c_i - \min_j c_j}{\max_j c_j - \min_j c_j},
\end{equation}
where $c_i$ denotes the raw confidence reported by reviewer $i$, and $\tilde{c}_i \in [\alpha,1]$ is the normalized confidence. Here, $\alpha \in [0,1]$ is a tunable parameter that controls the mapping of the lowest confidence level to the normalized scale. To address this question, we conducted experiments to evaluate which normalization strategies for confidence yield the most effective results.

As shown in Tab.~\ref{tab:RTS_param}, the smooth formulation of $\sigma(c)$ yields superior performance, indicating that gradual scaling of confidence is more effective than aggressive penalization. For the lower bound $\alpha$, the best results occur at $\alpha=0.2$, while both smaller and larger values reduce performance, showing that a moderate range is preferable.
\begin{table*}[ht]
\centering
\caption{Ablation study on RTS normalization strategies. 
(a) Different formulations of $\sigma(c)$. 
(b) Effect of the lower bound $\alpha$ for confidence normalization.}
\label{tab:RTS_param}
\begin{minipage}{0.46\linewidth}
\centering
\subcaption{Formulations of $\sigma(c)$}
\begin{tabular}{lccc}
\hline
$\boldsymbol{\sigma(c)}$ & \textbf{AUC} & \textbf{NDCG} & $\boldsymbol{\rho}$ \\
\hline
$\sigma_{\text{aggressive}}$ & 0.766 & 0.775 & 0.391 \\
$\sigma_{\text{smooth}}$     & \textbf{0.782} & \textbf{0.771} & \textbf{0.432} \\
\hline
\end{tabular}

\footnotesize
\emph{$\sigma_{\text{aggressive}}(c)=0.1+0.4(1-c)$,\;
$\sigma_{\text{smooth}}(c)=0.2(1-c)+0.05$.}
\end{minipage}
\hfill
\begin{minipage}{0.46\linewidth}
\centering
\subcaption{Confidence normalization (lower bound $\alpha$)}
\begin{tabular}{lccc}
\hline
\textbf{$\boldsymbol{\alpha}$} & \textbf{AUC} & \textbf{NDCG} & $\boldsymbol{\rho}$ \\
\hline
0.0 & 0.756 & 0.\textbf{796} & 0.391 \\
0.2 & \textbf{0.782} & 0.771 & \textbf{0.432} \\
0.5 & 0.764 & 0.772 & 0.405 \\
0.8 & 0.769 & 0.771 & 0.395 \\
\hline
\end{tabular}
\end{minipage}
\end{table*}

\section{Implementation of API-Only Baselines}
\label{app:API_details}

To ensure fair comparison and to examine the limitations of prompt-only approaches, we re-implemented baseline experiments using GPT-5-mini as an AI reviewer. 
In the \textit{pointwise} setting, the model was provided with the title and abstract of a single paper and instructed to output a normalized quality score in the range $[0,1]$. 
In the \textit{pairwise} setting, two papers were presented simultaneously, and the model was required to judge which paper exhibited higher overall quality. 
To further evaluate the robustness of prompt design, we constructed two distinct prompts for each setting, as reported in Tab.~\ref{tab:comparison_API_prompt}. 
We then evaluated the outputs using both ranking-based and classification-based metrics. 
For classification, the F1 score was computed under a decision threshold corresponding to the top-30\% predicted scores, following standard practice in acceptance prediction. 
This setup allows us to quantify the effectiveness of prompt-only methods and highlight their inherent weaknesses when compared with our proposed debiased pairwise learning framework.

\begin{table}[ht]
\renewcommand{\arraystretch}{1.5} 
\centering
\caption{Performance comparison.}
\label{tab:comparison_API_prompt}
\begin{tabular}{lp{6.5cm}cccc}
\hline
\textbf{Category} & \textbf{Prompt} & \textbf{F1}& \textbf{AUC}  & \textbf{NDCG} &   $\boldsymbol{\rho}$ \\
\hline
Pointwise   
    &  ``You are an expert academic paper reviewer.\textbackslash nEvaluate the quality of a single paper based on title and abstract only.\textbackslash n\textbackslash nConsider only:\textbackslash n- Novelty and significance\textbackslash n- Clarity of writing\textbackslash n- Technical soundness (as inferred from abstract)\textbackslash n\textbackslash nYou MUST output a STRICT JSON object ONLY with this exact schema:\textbackslash n\{\texttt{"score": 0.00}\}\textbackslash n\textbackslash nThe score must be a normalized quality between 0 and 1 (inclusive),\textbackslash nformatted with exactly two decimal places. Do NOT include any extra keys or text.'' 
    & 0.427 & 0.654 & 0.702 & 0.315 \\ 
\cline{2-6}
    & ``Review the paper from its title and abstract.\textbackslash nJudge quality by originality, clarity, and soundness.\textbackslash n\textbackslash nReturn ONLY a JSON object:\textbackslash n\{\texttt{"score": 0.00}\}\textbackslash n\textbackslash nScore $\in [0,1]$, exactly two decimals. No other text.'' 
    & 0.422 & 0.637 & 0.667 & 0.278 \\
\hline
Pairwise    
    &  ``You are an expert academic paper reviewer.\textbackslash nYour job is to decide which of two papers is of higher overall quality\textbackslash nbased on title and abstract only.\textbackslash n\textbackslash nConsider only:\textbackslash n- Novelty and significance\textbackslash n- Clarity of writing\textbackslash n- Technical soundness (as inferred from abstract)\textbackslash n\textbackslash nYou MUST choose exactly one better paper (no ties).\textbackslash nOutput a STRICT JSON object ONLY, with this exact schema:\textbackslash n\{\texttt{"better": a}\}  OR  \{\texttt{"better": b}\}\textbackslash n\textbackslash nDo NOT include any explanations, notes, or extra fields.\textbackslash nDo NOT include reasoning.\textbackslash nDo NOT include any text outside the JSON object.'' 
    & 0.448 & 0.655 & 0.686 & 0.297 \\
\cline{2-6}
    &  ``Decide which paper (a or b) is of higher quality\textbackslash nbased on title and abstract only. Consider novelty, clarity, and\textbackslash ntechnical soundness. Pick exactly one.\textbackslash n\textbackslash nOutput STRICT JSON:\textbackslash n\{\texttt{"better": a}\}  OR  \{\texttt{"better": b}\}\textbackslash n\textbackslash nNo extra text.'' 
    & 0.439 & 0.659 & 0.690 & 0.316 \\
\hline
\end{tabular}
\end{table}

\section{Bucket Distribution Definition}
\label{app:bucket_dist}

As noted in Section~\ref{sec:learn_difficulty}, we discretize the pairwise $RTS$ gap $\Delta_{ab}$ into buckets 
with edges $[0,0.1,0.2,\dots,1.0]$, which correspond to 9 intervals. We define five representative ratio distributions over these buckets to characterize different difficulty settings, 
ranging from \textit{Easiest} to \textit{Hardest}. 
Table~\ref{tab:bucket_ratios} lists the exact proportions. 
In particular, easier distributions allocate more weight to large gaps ($>0.8$), 
whereas harder ones emphasize small gaps ($<0.1$). 
The moderate case is close to uniform and serves as a neutral baseline.
\begin{table}[ht]
\centering
\caption{Bucket ratio definitions for different difficulty levels.}
\label{tab:bucket_ratios}
\resizebox{\textwidth}{!}{
\begin{tabular}{lccccccccc}
\hline
\textbf{Difficulty} & $<0.1$ & 0.1--0.2 & 0.2--0.3 & 0.3--0.4 & 0.4--0.5 & 0.5--0.6 & 0.6--0.7 & 0.7--0.8 & $>0.8$ \\
\hline
Easiest   & 0.03  & 0.04  & 0.06  & 0.08  & 0.10  & 0.12  & 0.14  & 0.16  & 0.27 \\
Easier    & 0.05  & 0.06  & 0.07  & 0.08  & 0.10  & 0.12  & 0.15  & 0.17  & 0.20 \\
Moderate  & 0.111 & 0.111 & 0.111 & 0.111 & 0.111 & 0.111 & 0.111 & 0.111 & 0.112 \\
Harder    & 0.20  & 0.17  & 0.15  & 0.13  & 0.10  & 0.08  & 0.07  & 0.06  & 0.04 \\
Hardest   & 0.27  & 0.16  & 0.14  & 0.12  & 0.10  & 0.08  & 0.06  & 0.04  & 0.03 \\
\hline
\end{tabular}}
\end{table}

\section{Data Engine for Constructing NAIDv2} 
\label{app:NAIDv2 Engine}

To enable scalable data collection and management, we implemented a dedicated data engine for NAIDv2 (Tab.~\ref{tab:naidv2_schema}). 
The engine adopts a three-layer architecture: the Controller layer interfaces with external APIs (e.g., OpenReview) and orchestrates data flow; 
the Service layer parses raw submissions and reviews, normalizes metadata (such as confidence scores), and exports processed datasets; 
and the DAO layer provides object-oriented access to the underlying MySQL database.

The database consists of two core tables, `Papers' and `Reviews', linked by the paper identifier. 
The `Papers' table stores submission-level metadata (title, abstract, venue, derived GT labels, etc.), 
while the `Reviews' table stores reviewer comments, ratings, and rebuttals. 
Automatic crawlers retrieve all submissions from OpenReview, insert missing entries, and populate reviews through structured parsing. 

This modular design ensures reproducibility, facilitates large-scale updates across multiple years of OpenReview data, 
and supports downstream model training and evaluation.

\begin{table*}[t]
\centering
\caption{Schema of the Papers and Reviews tables in NAIDv2 (partial view). PK = Primary Key, FK = Foreign Key.}

\label{tab:naidv2_schema}
\scriptsize
\begin{minipage}{0.47\linewidth}
\centering
\subcaption{Papers Table}
\resizebox{\linewidth}{!}{%
\begin{tabular}{lll}
\hline
\textbf{Field} & \textbf{Type} & \textbf{Desc.} \\
\hline
\underline{id} & BIGINT & PK \\
title & VARCHAR(255) & Title \\
abstract & TEXT & Abstract \\
external\_id & VARCHAR(255) & Openreview ID \\
openreview\_venue & VARCHAR(255) & Venue \\
scores & JSON & Review scores \\
confs & JSON & Confidences \\
pub\_year & INT & Year \\
embedding & JSON & Embedding vec \\
cluster\_cat & VARCHAR(64) & Cluster label \\
\hline
\end{tabular}}
\end{minipage}%
\hfill
\begin{minipage}{0.47\linewidth}
\centering
\subcaption{Reviews Table}
\resizebox{\linewidth}{!}{%
\begin{tabular}{lll}
\hline
\textbf{Field} & \textbf{Type} & \textbf{Desc.} \\
\hline
\underline{id} & BIGINT & PK \\
\textit{paper\_id} & BIGINT & FK $\to$ Papers.id \\
belong\_to & VARCHAR(128) & Group ID \\
reply\_to & VARCHAR(128) & Reply target \\
conf\_score & FLOAT & Confidence \\
rec\_score & FLOAT & Recommendation \\
reply\_title & TEXT & Reply title \\
review\_summary & TEXT & Summary \\
strengths & TEXT & Strengths \\
weakness & TEXT & Weaknesses \\
\hline
\end{tabular}}
\end{minipage}
\end{table*}

\section{Effect of Additional Information}
\label{app:additional_info}
As noted in NAIPv1~\citep{zhao2025words}, incorporating additional information can sometimes enhance model performance, though such gains are not always consistent. In this study, adhering to the double-blind requirement, we extracted auxiliary information directly from the papers and incorporated it into network training. Specifically, we employed a computer vision-based approach to extract key figures and key tables: the scores for the key figures were derived through network inference, and the table descriptions were generated using analysis by GPT-5-mini. However, our experiments show that adding such auxiliary content does not improve performance across the tested settings. In fact, models augmented with introductions, conclusions, figure scores, or table descriptions all perform worse than those relying solely on titles and abstracts. These results are broadly consistent with prior findings~\citep{zhou2024llm}, indicating that not all types of supplementary content provide useful signals for review score prediction, further confirming that abstracts already contain sufficient information.

\begin{table}[ht]
\centering
\caption{Comparison of different model architectures for review score prediction.}
\label{tab:additional_info}
\begin{tabular}{lccc}
\hline
\textbf{Method} & \textbf{AUC} & \textbf{NDCG} & $\boldsymbol{\rho}$ \\
\hline
Title \& Abstract (\textit{ours}) & \textbf{0.782}     & 0.771     &\textbf{ 0.432} \\
+Introdution           & 0.716     & 0.760     & 0.321 \\
+Conclusion        & 0.750     & 0.768  & 0.395\\
+Key Figure Score      & 0.769     & \textbf{0.798}     & 0.417 \\
+Key Table Description  & 0.758     & 0.753     & 0.381 \\  
\hline
\end{tabular}
\end{table}

\section{Additional Remarks on Experimental Comparisons}
\label{app:special_note}
\begin{figure}[h]
\centering
\begin{minipage}[b]{0.45\textwidth}
\centering
\includegraphics[width=\linewidth]{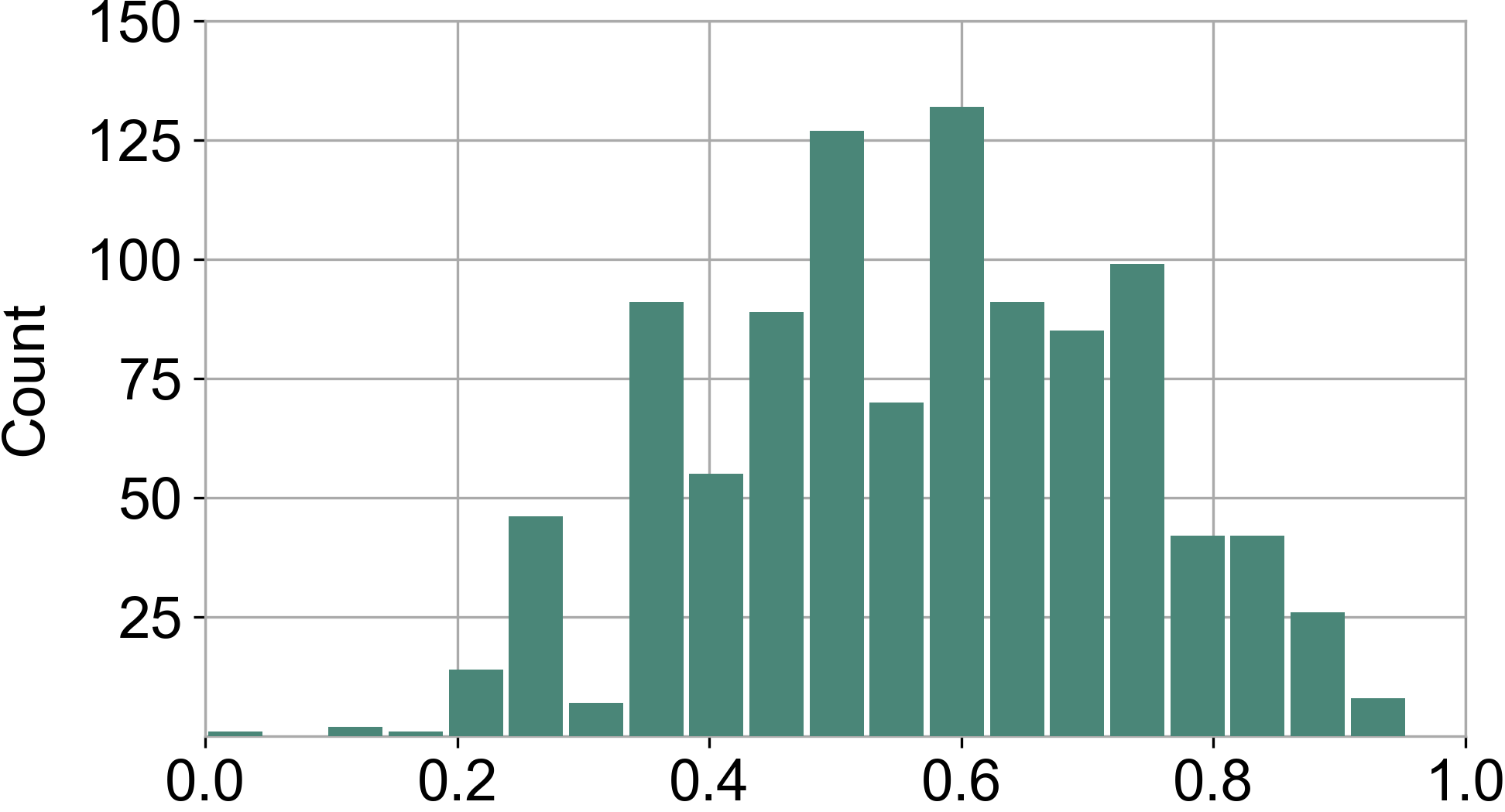}
\caption*{(a) NAIDv2 (\textit{Ours})}
\end{minipage}
\hspace{2mm}
\begin{minipage}[b]{0.45\textwidth}
\centering
\includegraphics[width=\linewidth]{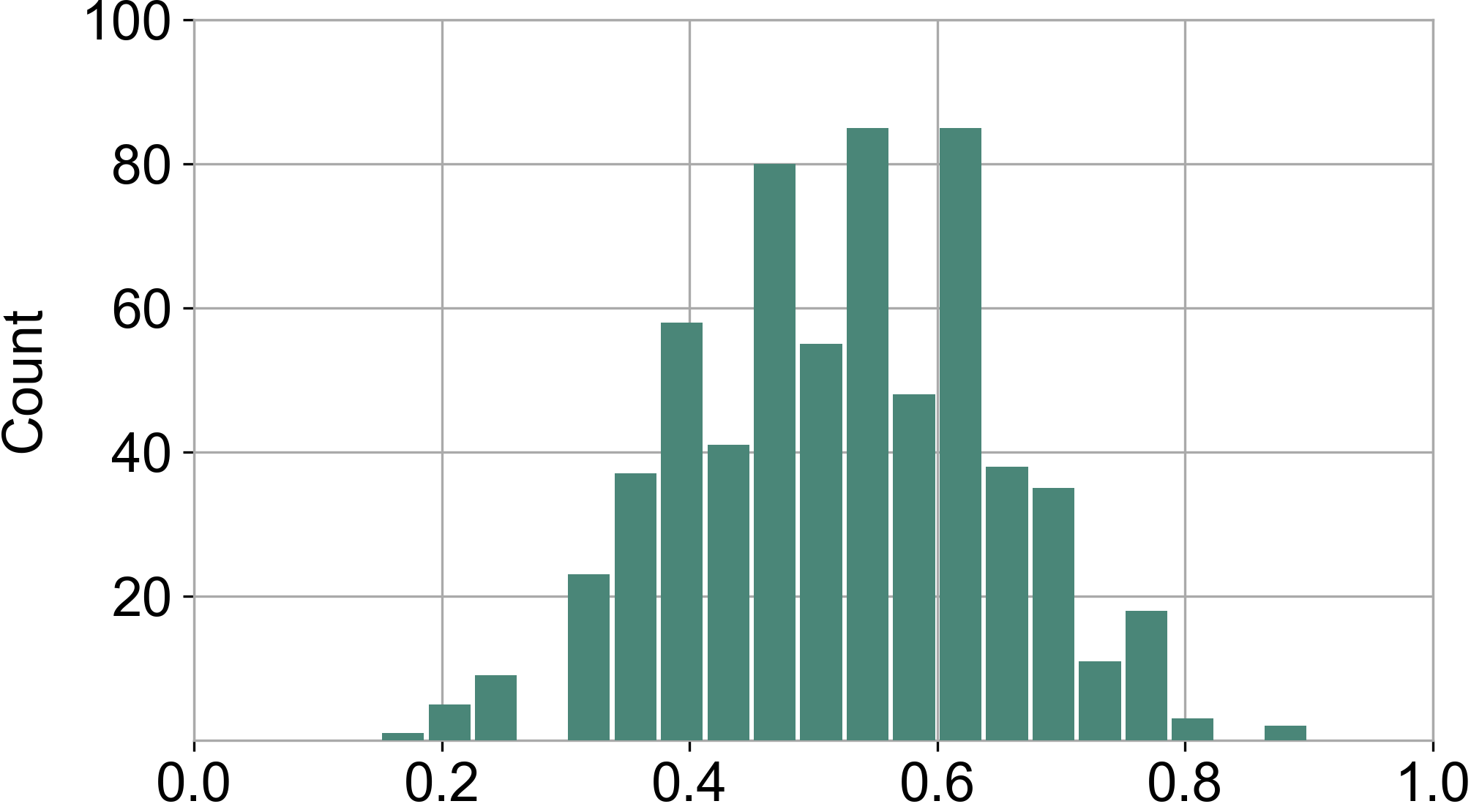}
\caption*{(b) DeepReviewer}
\end{minipage}

\caption{\textbf{Data distribution comparison between NAIDv2 (ours) and DeepReviewer Test 2025 datasets.} The two distributions are well aligned, with a Jensen–Shannon of 0.079, indicating comparability of evaluation results across methods.}
\label{fig:data-distribution-comparison}
\end{figure}

In Tab.~\ref{tab:comparison_sota}, we compare the performance of existing methods, reporting some results directly from the original publications. We were unable to reproduce SEA-E~\citep{yu2024automated}, OpenReviewer~\citep{idahl2024openreviewer}, DeepReviewer~\cite{zhu-2025-deepreview}, and CycleReviewer~\citep{weng2025cycleresearcher} for two primary reasons. First, the training and test sets used in these works differ and sometimes overlap, raising the possibility that our test data appear in their training sets (and vice versa). Second, retraining these models is computationally infeasible. For example, CycleReviewer requires training a 123B-parameter model, which far exceeds our available resources. Moreover, we note that instances from benchmark datasets may be embedded in the training data of these methods, suggesting that their reported gains could partly reflect data leakage rather than genuine improvements in generalization.

Nonetheless, most prior studies follow a relatively consistent dataset construction protocol, typically sampling randomly from the ICLR dataset. As shown in Fig.~\ref{fig:data-distribution-comparison}, the distribution of our test set is well aligned with that of DeepReviewer, with a Jensen–Shannon of 0.079. This alignment indicates that, despite the aforementioned limitations, the comparisons in Tab.~\ref{tab:comparison_sota} remain reasonably meaningful across methods.

\section{Further Details on Embedding and Clustering}

A key motivation for our embedding–clustering design arises from the observation that review scores can vary substantially across domains. However, directly applying keyword-based normalization often suffers from issues of uncontrollable granularity and drifting category boundaries, which weaken its reliability. For instance, research on remote sensing~\citep{li2024sm3det} may be regarded either as satellite-based sensing~\citep{ali2025advanced,kuckreja2024geochat} or as visual perception~\citep{girshick2015fast,li2025vitp,zhang2025rsar}. ``PromptKD''~\citep{li2024promptkd} might be situated within prompt learning~\citep{li2024advancing,wu2024cascade,chen2024revisiting} or knowledge distillation~\citep{hinton2015distilling,li2023curriculum}.  To address this challenge, we employ Qwen3-Embedding in combination with hierarchical clustering, which enables semantically grounded grouping of papers beyond brittle keyword matching.

As shown in Table~\ref{tab:embed_cluster} and Table~\ref{tab:qwen3_prompt}, this setup allows us to systematically study the impact of embedding and clustering choices. Both BGE-M3~\citep{chen2024bge} and Qwen3-Embedding provide strong baselines, yet Qwen3-Embedding paired with hierarchical clustering achieves the most consistent gains. HDBSCAN~\cite{mcinnes2017hdbscan} remains competitive but trails slightly in AUC and $\rho$. Importantly, embeddings enriched with explicit clustering-oriented instructions outperform those without, and incorporating both titles and abstracts yields better representations than titles alone. Among the prompts we explored, the one explicitly directing the model to group semantically similar papers provides the most balanced improvements across all evaluation metrics.

\begin{table*}[ht]
\centering
\caption{Impact of embedding models and clustering algorithms on retrieval performance. 
(a) Comparison of embedding models. 
(b) Comparison of clustering algorithms with confidence normalization.}
\label{tab:embed_cluster}
\begin{minipage}{0.46\linewidth}
\centering
\subcaption{Embedding models}
\begin{tabular}{lccc}
\hline
Model & \textbf{AUC↑} & \textbf{NDCG↑} & $\boldsymbol{\rho}$ \\
\hline
BGE-M3 & 0.774 & 0.769 & 0.417 \\
Qwen3-emb & \textbf{0.782} & \textbf{0.771} & \textbf{0.432} \\
\hline
\end{tabular}
\end{minipage}
\hfill
\begin{minipage}{0.46\linewidth}
\centering
\subcaption{Clustering algorithms}
\begin{tabular}{lccc}
\hline
Method & \textbf{AUC↑} & \textbf{NDCG↑} & $\boldsymbol{\rho}$ \\
\hline
HDBSCAN & 0.765 & \textbf{0.803} & 0.422 \\
Hier. Cluster & \textbf{0.782} & 0.771 & \textbf{0.432} \\
\hline
\end{tabular}
\end{minipage}
\end{table*}

\begin{table}[ht]
\centering
\caption{Comparison of different instruction prompt designs}
\label{tab:qwen3_prompt}
\begin{tabular}{p{6cm}ccc}
\hline
\textbf{Instruction} & \textbf{AUC↑} & \textbf{NDCG↑} & $\boldsymbol{\rho}$ \\
\hline
None \textit{(no instruction)} & 0.759 & 0.775 & 0.266 \\
Identify the main category of scholar papers based on the titles and abstracts & 0.767 & \textbf{0.831} & 0.416 \\
Given a title of a scientific paper, retrieve the titles of other relevant papers & 0.768 & 0.784 & 0.412 \\
Cluster similar papers as close as possible based on title and abstract (\textit{ours}) & \textbf{0.782 }& 0.771 & \textbf{0.432} \\
\hline
\end{tabular}
\end{table}

\section{Ablation on Pairwise Sampling Strategies}
\label{app:pairwise_strategy}
To further investigate the robustness of the proposed pairwise learning strategy, we conduct ablation studies by introducing additional constraints during pair construction. The first experiment (Table~\ref{tab:pairwise_constraints} Panel (a)) examines the effect of limiting the maximum number of times a sample can appear in training pairs. The results show that imposing such restrictions does not provide clear benefits. In fact, the best overall performance is achieved when no limitation is enforced, suggesting that allowing samples to be paired freely provides greater diversity and stronger supervision signals for training.

The second experiment (Table~\ref{tab:pairwise_constraints} Panel (b)) analyzes the role of enforcing a minimum score difference between paired samples. Without such a margin, performance degrades, indicating that constructing pairs from nearly indistinguishable items introduces noise into the training signal. A margin of 0.05 provides the best overall balance, improving both AUC and Spearman correlation, while larger margins ($\geq 0.1$) result in performance drops due to reduced pair availability. These findings highlight the importance of carefully controlling pair selection to enhance both the reliability and effectiveness of pairwise learning.

\begin{table*}[h]
\centering
\caption{Effect of pairwise learning constraints. 
(a) Influence of limiting the maximum occurrence times of each sample in pairwise training. 
(b) Influence of enforcing a minimum score difference when constructing pairs.}
\label{tab:pairwise_constraints}
\begin{minipage}{0.46\linewidth}
\centering
\subcaption{Maximum occurrence times}
\begin{tabular}{lccc}
\hline
\textbf{Setting} & \textbf{AUC} & \textbf{NDCG} & $\boldsymbol{\rho}$ \\
\hline
2  & 0.776 & 0.773 & 0.426 \\
4  & 0.766 & 0.791 & 0.408 \\
8  & 0.767 & 0.774 & 0.391 \\
16 & 0.782 & 0.771 & 0.432 \\
32 & \textbf{0.782} & \textbf{0.771} & \textbf{0.432} \\
\hline
\end{tabular}
\end{minipage}
\hfill
\begin{minipage}{0.46\linewidth}
\centering
\subcaption{Minimum score difference}
\begin{tabular}{lccc}
\hline
\textbf{Setting} & \textbf{AUC} & \textbf{NDCG} & $\boldsymbol{\rho}$ \\
\hline
0    & 0.752   & 0.778   & 0.395   \\
0.05 & \textbf{0.782} & 0.771 & \textbf{0.432} \\
0.1  & 0.753 & \textbf{0.785} & 0.397 \\
0.15 & 0.770 & 0.778 & 0.402 \\
0.2  & 0.761  & 0.774 & 0.404 \\
\hline
\end{tabular}
\end{minipage}
\end{table*}




\section{Discussion and Practical Considerations}
\label{sec:disscussion}
\begin{wraptable}[6]{r}{0.4\linewidth}
\centering
\caption{Performance of NAIPv1/v2 on article impact prediction task.}
\label{tab:eval_on_v1}
\begin{tabular}{lcc}
\hline
\textbf{Model} & \textbf{NDCG} & \boldsymbol{$\rho$} \\
\hline
NAIPv1 & \textbf{0.901} & \textbf{0.459} \\
NAIPv2 & 0.761 & 0.279 \\
\hline
\end{tabular}
\end{wraptable}

As shown in Tab.~\ref{tab:eval_on_v1}, NAIPv1 performs strongly in predicting the academic impact of newly published articles, whereas NAIPv2 underperforms on the NAIDv1 benchmark. This result appears consistent with the observation of ~\cite{cortes2021inconsistency} that impact and quality are often only weakly correlated.

These findings suggest that there is no single ``universal key'' for all tasks, and the choice of method should depend on the specific application. API-based solutions are easy to use and broadly accessible, but their performance is limited (Tab.~\ref{tab:comparison_sota}); fine-tuned autoregressive models benefit from domain-specific optimization and achieve stronger results, though at the cost of additional deployment requirements; NAIPv2 provides performance comparable to these autoregressive approaches with much faster inference, but lacks the interpretive outputs that autoregressive models offer.

Overall, these comparisons indicate that no single method prevails across all scenarios, underscoring the importance for both research and industry communities to recognize and leverage the complementary strengths of these approaches.

\end{document}